\documentclass[letterpaper]{article} 
\usepackage{aaai25}  
\usepackage{times}  
\usepackage{helvet}  
\usepackage{courier}  
\usepackage[hyphens]{url}  
\usepackage{graphicx} 
\usepackage{natbib}  
\usepackage{caption} 
\usepackage[labelformat=simple]{subcaption}

\usepackage{enumitem}
\usepackage{booktabs}

\usepackage{algorithm}
\usepackage{algorithmic}
\usepackage{enumitem}

\usepackage{tcolorbox}
\usepackage{array}

\usepackage{amsmath}
\usepackage{tikz}
\usepackage{pgf}
\usepackage{tabularx}
\usepackage{amssymb}
\usepackage{subcaption}

\usepackage{newfloat}
\usepackage{listings}
\DeclareCaptionStyle{ruled}{labelfont=normalfont,labelsep=colon,strut=off} 
\floatstyle{ruled}
\newfloat{listing}{tb}{lst}{}
\floatname{listing}{Listing}
\begin{document}

\title{LEAF: Learning and Evaluation Augmented by Fact-Checking to Improve Factualness in Large Language Models}

\author{Hieu Tran, $^1,^2$, Junda Wang  $^1,^2$, Yujan Ting  $^2$, Weijing Huang$^{2}$, Terrence Chen  $^2$ \\
    $^1$ United Imaging Intelligence, MA, USA \\
$^2$ Manning College of Information and Computer Sciences, University of Massachusetts Amherst, MA, USA\\
  {\tt \{hieutran, jundawang\}@umass.edu}\\ 
  {\tt \{yujan.ting, weijing.huang, terrance.chen\}@uii-ai.com}\\ 
}
\maketitle

\begin{abstract}
Large language models (LLMs) have shown remarkable capabilities in various natural language processing tasks, yet they often struggle with maintaining factual accuracy, particularly in knowledge-intensive domains like healthcare. This study introduces LEAF: Learning and Evaluation Augmented by Fact-Checking, a novel approach designed to enhance the factual reliability of LLMs, with a focus on medical question answering (QA). LEAF utilizes a dual strategy to enhance the factual accuracy of responses from models such as Llama 3 70B Instruct and Llama 3 8B Instruct. The first strategy, Fact-Check-Then-RAG, improves Retrieval-Augmented Generation (RAG) by incorporating fact-checking results to guide the retrieval process without updating model parameters. The second strategy, Learning from Fact-Checks via Self-Training, involves supervised fine-tuning (SFT) on fact-checked responses or applying Simple Preference Optimization (SimPO) with fact-checking as a ranking mechanism, both updating LLM parameters from supervision. Experimental results demonstrate that LEAF not only effectively detects inaccurate responses but also significantly enhances the model's accuracy. These findings suggest that integrating fact-checked responses—whether through RAG enhancement or self-training—enhances the reliability and factual correctness of LLM outputs, offering a promising solution for applications where information accuracy is crucial.
\end{abstract}




\begin{figure*}[ht!]
\centering
\includegraphics[width=\textwidth]{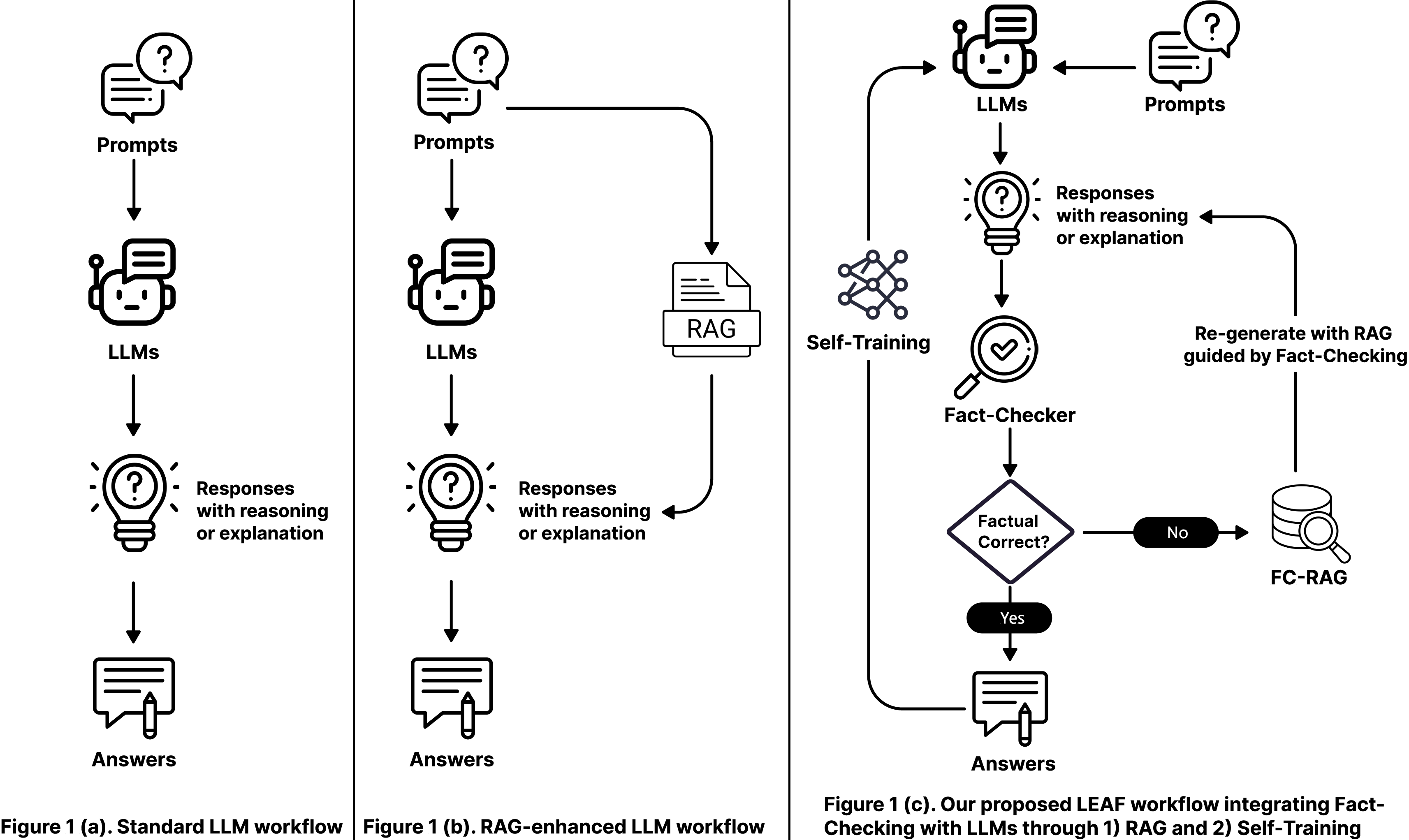}
\caption{Comparison of workflows: standard LLM workflow (left), RAG-enhanced LLM workflow (middle), and our proposed Fact-Checking integrated workflow (right).}
\label{fig:combined}
\end{figure*}


\section{Introduction}

Large language models (LLMs) have revolutionized natural language processing (NLP), demonstrating remarkable capabilities in generating coherent and contextually relevant text. However, a significant challenge persists across various applications: LLMs sometimes generate plausible yet factually incorrect or unverified content~\citep{ji2023survey,bang2023multitask}. This issue is particularly concerning in domains such as healthcare, where the accuracy and reliability of information are critical.

The issue of factual inconsistency in LLMs occurs due to various factors, including insufficient training data, biases in the training corpus, and the model's inherent limitations in understanding complex real-world knowledge. The tendency to produce content that seems plausible but may not be factually grounded can potentially cause harm if used in sensitive contexts like medical diagnosis or treatment recommendations, underscoring the need for effective mitigation strategies \cite{zellers2019defending, marcus2020gpt}.

In order to address this challenge, fact-checking has emerged as a promising approach. Fact-checking mechanisms involve verifying the factual accuracy of generated content against reliable data sources, thus providing a filter to identify and rectify misinformation. Prior work has investigated a variety of methods for integrating fact-checking into LLM workflows, including retrieval-augmented generation (RAG) and other verification techniques such as Factcheck-GPT~\cite{lewis2020retrieval, petroni2021kilt, wang2023factcheck}. However, existing approaches have notable limitations. Proprietary models like Factcheck-GPT cannot be deployed on private datasets, restricting their use in sensitive domains such as healthcare. Moreover, these models are often designed for general-purpose use rather than specialized for specific areas like medical expertise. Their inability to be fine-tuned further limits their effectiveness in specialized domains that require nuanced understanding.

In this study, we introduce LEAF: Learning and Evaluation Augmented by Fact-Checking, a novel approach featuring two parallel strategies for leveraging LLMs. The first strategy enhances LLM applications without updating model parameters, which is particularly suitable for proprietary models like ChatGPT that can't be fine-tuned. The second strategy employs self-training to update LLM parameters, enabling the model to exploit extensive unlabeled data by iteratively refining its performance using its own predictions combined with fact-checking results as pseudo-labels. These strategies represent our two main contributions:

\begin{itemize}
    \item \textbf{Fact-Check-Then-RAG}: We propose a novel approach to Retrieval-Augmented Generation (RAG) where the retrieval process is guided by the results of fact-checking. This method ensures that the retrieved information specifically enhances the factual accuracy in the model's initial output, leading to more contextually relevant and accurate responses without updating the underlying LLM parameters.
    
    \item \textbf{Learning from Fact-Check via Self-Training}: We explore self-training mechanisms using fact-checked responses to improve factualness, which encompasses two parallel approaches that update LLM parameters: a) Supervised Fine-Tuning (SFT) on fact-checked responses: We implement a process of generating multiple responses to a given query, evaluating these responses for factual accuracy using a fact-checking system, and then fine-tuning the model based on the responses that pass the fact-check. b) Simple Preference Optimization (SimPO)~\cite{meng2024simpo} using fact-checking as a ranking mechanism: We employ the fact-checking system to rank generated responses, selecting the highest-scoring ones as ``chosen" and the lowest-scoring ones as ``rejected" for optimization. We demonstrate that our model maintains effectiveness across two different training methods, showcasing its robustness in learning from fact-checking. 
\end{itemize}

This self-training approach within LEAF, using either SFT or SimPO, is particularly valuable in low-resource domains, where labeled data is scarce, as it leverages the model's capability to self-improve by continuously refining its knowledge through validated responses. Our findings show that both methods can significantly enhance model performance, offering new pathways for learning without the need for labeled data. By integrating these strategies, LEAF provides a robust framework for improving LLMs' performance, enhancing their reliability and factual correctness, particularly in critical domains like healthcare where the accuracy of information is important.

\section{Methodology}
In this section, we describe our proposed methodology to enhance the factual accuracy and reliability of large language models in generating responses. Our approach, LEAF, integrates fact-checking, retrieval-augmented generation, and self-training mechanisms to systematically improve factuality in LLM outputs. The workflow of our proposed method is illustrated in Figure~\ref{fig:combined}.


The proposed workflow is developed to enhance the factual accuracy of LLM-generated responses by integrating a rigorous fact-checking process. In the conventional LLM workflow (Figure~\ref{fig:combined}(a)), the model generates responses to prompts, providing reasoning or explanations, and directly delivers the final answers. However, this approach does not inherently ensure the factual correctness of the generated content. To address this limitation, we introduce LEAF Mechanism I: Fact-Check-Then-RAG. As depicted in our enhanced workflow (Figure~\ref{fig:combined}(c)), after the LLM generates a response, it undergoes evaluation by a fact-checking system. If the response is deemed factually accurate, it is retained as the final output. Conversely, if the response is identified as incorrect, the workflow triggers a Retrieval-Augmented Generation (RAG) approach, as illustrated in Figure~\ref{fig:combined}(b). In this phase, relevant documents retrieved during the fact-checking process are integrated into the prompt, guiding the model to regenerate a more accurate response. This iterative process continues until a factually correct answer is achieved. Additionally, we propose LEAF Mechanism II, wherein factually verified responses are employed for self-training. The model is fine-tuned on these fact-checked outputs, thereby further enhancing its performance and accuracy in producing reliable and factual responses. The subsequent subsections provide detailed descriptions of our methodology components.

\subsection{Fact-Checking for LLM Responses}

We leverage the Search-Augmented Factuality Evaluator (SAFE) \cite{wei2024long}, adapting it specifically for the medical domain to evaluate the factual accuracy of LLM-generated responses. SAFE overcomes the limitations of traditional evaluation methods that rely on preset reference answers, which are often insufficient for complex, long-form responses. SAFE's design includes breaking down responses into individual facts and dynamically verifying these facts through iterative Google Search queries, ensuring precise and timely evaluations.

For our adaptation, we focused on the following enhancements to make the system zero-cost, controllable, and effective, especially in the medical domain:

\begin{itemize}
\item \textbf{Incorporation of Question Context}: In medical QA tasks, it is crucial to consider the context provided by the question, as these often involve specific scenarios that require accurate and nuanced responses. Our adaptation ensures that the fact-checking process integrates this context, leading to more accurate assessments of the LLM responses' relevance and correctness. More detailed information on the prompts used in this process can be found in the Appendix table \ref{tab:prompts}.

    \item \textbf{Deployment of Qwen2-72B-Instruct as the Rater}: To enhance the reliability of the factuality ratings, we replaced GPT-3.5 with the Qwen2-72B-Instruct \cite{yang2024qwen2} large language model. This model offers advanced capabilities in evaluating complex medical facts and ensuring that the responses are both accurate and relevant.

    \item \textbf{Use of MedRAG Corpus with ColBERT Retrieval}: Instead of using Google search results, which may not always be relevant or accessible for medical queries, we employ the MedRAG corpus \cite{xiong2024benchmarking}. This corpus includes authoritative sources such as Wikipedia, PubMed, textbooks, and StatPearls. By using the ColBERT \cite{khattab2020colbert} retrieval model, we can efficiently extract relevant documents from these sources. This approach ensures that our fact-checking system remains zero-cost, fully controllable, and tailored to the specific needs of the medical domain.

\end{itemize}

Our comparative analysis shows that this adapted system outperforms Factcheck-GPT~\cite{wang2023factcheck}—a similar system that uses ChatGPT-3.5 and Google search—in terms of filtered accuracy in medical QA tasks. This indicates that our approach is more adept at ensuring factual correctness in the responses generated by LLMs, providing a robust and scalable solution for verifying the accuracy of information in critical domains like healthcare.

\subsection{Calculation of Fact-Check Scores}
To evaluate the factual accuracy of generated responses, we employed a sentence-level fact-checking approach. Each response is decomposed into individual sentences, and each sentence is independently verified against retrieved external knowledge sources. The fact-checking system assesses whether each sentence is supported by the retrieved knowledge. Specifically, for each sentence in the response, the system attempts to retrieve relevant documents or facts that confirm or refute the content of the sentence. A sentence is considered supported if the retrieved knowledge substantiates its factual accuracy. 

The fact-check score for a response is calculated as the ratio of supported sentences to the total number of sentences in the response. Formally, the fact-check score \( LEAF \) for a response is given by:
\[
LEAF = \frac{\text{Number of Supported Sentences}}{\text{Total Number of Sentences in the Response}}
\]


\begin{figure*}[h!]
    \centering
     \resizebox{1.0\textwidth}{!}{
    \begin{tikzpicture}[font=\small]

        \node[draw, thick, rounded corners, fill=green!20, text width=8cm, align=left, minimum height=4cm] (safe) at (0,0) {
            \textbf{Fact-Check} \\
            \vspace{3mm}
            \textbf{C) Hyperstabilization of microtubules} 
             \textbf{Not Supported} \textcolor{red}{\(\times\)} \\
            \vspace{0.3cm}
            \begin{tabularx}{\linewidth}{lX}
                \textbf{Search query \#1:} & What drugs are used to treat transitional cell carcinoma of the bladder that cause sensorineural hearing loss? \\
                \textbf{Result:} & This has many causes. The common high-frequency sensorineural type of hearing loss \dots  \\
                \textbf{Search query \#2:} & cisplatin side effects \\
                \textbf{Result:} & Chemotherapeutic Agents Cisplatin and carboplatin are accumulated by proximal tubular cells \dots \\
                \textbf{Search query \#3:} & What chemotherapy agent for bladder cancer causes hearing loss due to hyperstabilization of microtubules? \\
                \textbf{Result:} & Cytotoxic Chemotherapy Agents Table 103e-4 lists commonly used cytotoxic cancer chemotherapy agents \dots \\
                \textbf{Final reasoning:} & Given that cisplatin is a common chemotherapy for bladder cancer, and considering the mechanisms of action of the options provided, option C) Hyperstabilization of microtubules is not the most plausible choice. 
            \end{tabularx}
        };

        \node[draw, thick, rounded corners, fill=blue!20, text width=8cm, align=left, minimum height=8cm] (rag) at (10,0) {
            \textbf{RAG} \\
            \vspace{3mm}
         Given a multiple choice question, please select the correct answer and also provide a detailed reasoning for your choice. You can using the information provided in the knowledge section if necessary. \\
            \vspace{3mm}
            \textbf{Knowledge:} 
            \begin{tabularx}{\linewidth}{lX}
                \textbf{Search result \#1:} & This has many causes. The common high-frequency sensorineural type of hearing loss \dots  \\
                \textbf{Search result \#2:} & Chemotherapeutic Agents Cisplatin and carboplatin are accumulated by proximal tubular cells \dots \\
                \textbf{Search result \#3:} & Cytotoxic Chemotherapy Agents Table 103e-4 lists commonly used cytotoxic cancer chemotherapy agents \dots \\
            \end{tabularx}
            \textbf{Question:} A 67-year-old man with transitional cell carcinoma of the bladder comes to \dots The expected beneficial effect of the drug that caused this patient's symptoms is most likely due to which of the following actions? \\
            \textbf{Options:} \\
            (A) Inhibition of thymidine synthesis \\
            (B) Inhibition of proteasome \\
            (C) Hyperstabilization of microtubules \\
            (D) Generation of free radicals \\
            (E) Cross-linking of DNA \\
            \vspace{3mm}
            \textbf{Answer:} 
            \textbf{E) Cross-linking of DNA} 
            \textbf{Supported} \(\checkmark\)
        };
            \vspace{0.3cm}

        \draw[->, thick] (safe.east) -- node[above, midway] {then} (rag.west);

    \end{tikzpicture}}
    \caption{Fact-Check-Then-RAG is able to change the answer of LLMs by leveraging the knowledge retrieved from fact-check stage to regenerate the responses.}
    \label{fig:fcrag}
\end{figure*}
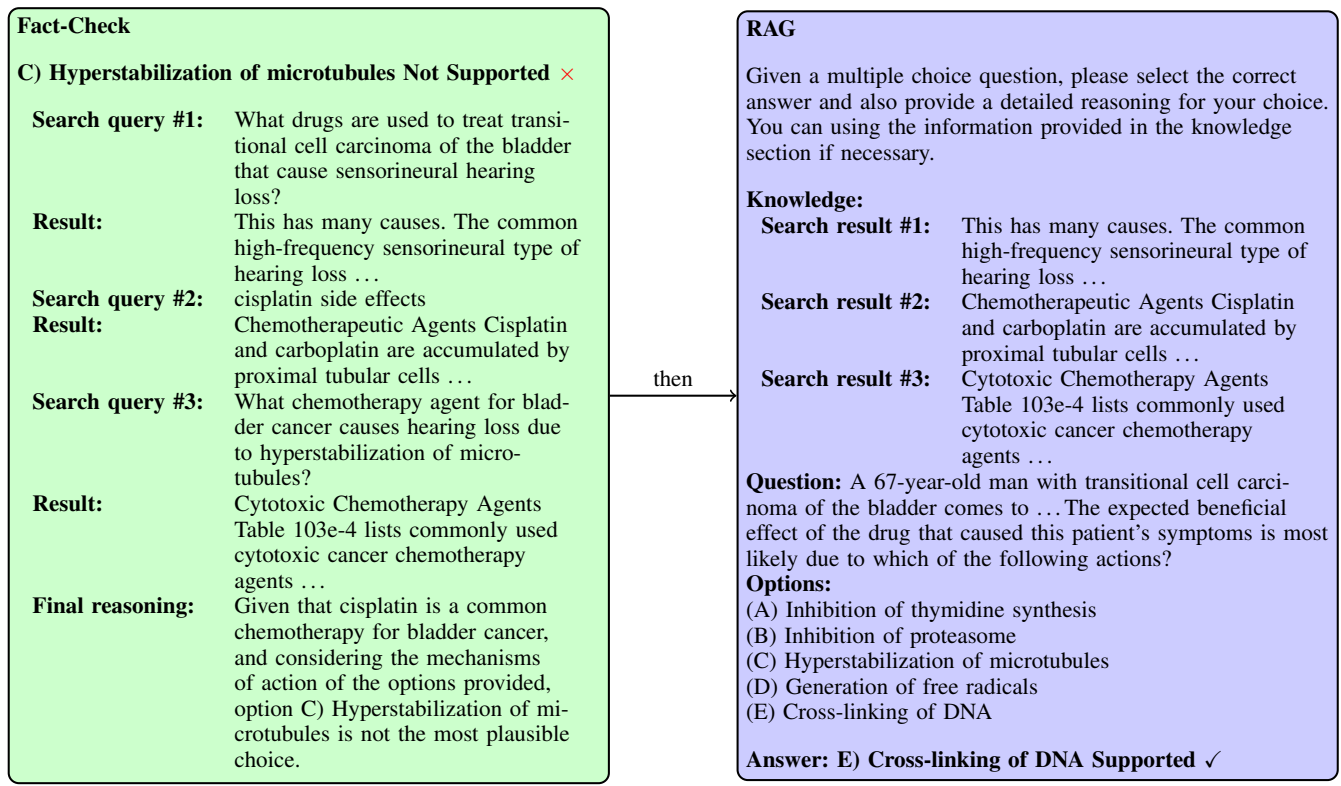


\subsection{LEAF Mechanism I: Fact-Check-Then-RAG}
Our first innovative mechanism, Fact-Check-Then-RAG, seamlessly integrates the fact-checking stage with Retrieval-Augmented Generation (RAG). This approach leverages the documents retrieved during the fact-checking process to enhance the generation of responses. The key idea is to utilize the knowledge retrieved from the fact-checking stage, specifically for individual facts that did not pass the fact-check test. This strategy ensures that when a fact is not supported by the retrieved knowledge sources, the relevant documents are included in the RAG prompt to help the LLM refine its reasoning or answer, potentially improving performance. As illustrated in Figure \ref{fig:fcrag}, the methodology involves several steps:

First, during the fact-checking stage, each individual fact in a response is evaluated for factual correctness using SAFE. If a fact is not supported by the knowledge retrieved (i.e., it fails the fact-check test), it indicates a gap between the LLM and the knowledge base. For these unsupported facts, relevant documents are retrieved from a comprehensive medical corpus (MedRAG), which includes authoritative sources like Wikipedia, PubMed, textbooks, and StatPearls. The ColBERT retrieval model is used to extract these documents. 

Next, the retrieved documents are included in the RAG prompt. This additional context provides the LLM with the necessary information to adjust its reasoning or answer, addressing the knowledge gap identified during the fact-checking stage. The LLM then generates new responses using the RAG framework, which is now enhanced with the relevant knowledge retrieved earlier. This iterative process ensures that the LLM's output is more informed and accurate.

By integrating fact-checking with RAG, our approach effectively addresses the knowledge gaps identified during the fact-checking process. This method enhances the LLM's ability to produce accurate and reliable responses, demonstrating improved performance over traditional RAG methods, particularly in increasing the factualness of generated content.


\subsection{LEAF Mechanism II: Learning from Fact-Check via Self-Training} 
We explore self-training mechanisms using fact-checked responses to enhance the performance of LLMs. This approach consists of two main parts: supervised fine-tuning on factually correct responses and optimization with Simple Preference Optimization.

\subsubsection{Supervised Fine-Tuning on Factually Correct Responses}

The first part involves fine-tuning the model using responses that have passed the fact-check test. This ensures the model is trained on verified, accurate information, thereby improving its overall performance. The process is as follows:
\begin{itemize}
    \item Response Generation and Fact-Checking: The LLM generates multiple responses to a given prompt, which are then evaluated using the fact-checking system.
    \item Selection of Factual Responses: Only those responses that pass the fact-checking process are selected for fine-tuning. And  "pass" is defined as the \textit{LEAF} score of the response is 1. 
    \item Fine-Tuning: The model is fine-tuned on these factually correct responses, reinforcing its ability to produce accurate and reliable outputs.
\end{itemize}

\subsubsection{Optimization with SimPO}

The second part of our self-training approach utilizes Simple Preference Optimization \cite{meng2024simpo} to rank and optimize responses based on their factual accuracy. SimPO aligns the reward formulation directly with the generation metric, eliminating the need for a reference model. This process involves Fact-Checking as a Ranking Model: The fact-checking system assigns scores to generated responses based on their factual accuracy. The highest-scoring responses are selected as ``chosen" and the lowest-scoring ones as ``rejected." By using the fact-checking system as a ranking model, SimPO effectively guides the model to prefer factually accurate responses.

\section{Experiments}
We detail two main experimental settings across different model configurations. For the Llama 3 70B Instruct model, we implemented some techniques to improve the performance of the model without updating the model parameters. In contrast, with the Llama 3 8B Instruct model, we explored self-training techniques where the parameters were updated based on fact-checking rather than labeled data. The self-training was performed using either supervised fine-tuning or SimPO, with the training data curated through a rigorous fact-checking process. Overall, while the Llama 3 70B Instruct model’s parameters remained fixed, the Llama 3 8B Instruct model was dynamically improved by learning directly from fact-check outcomes.
\subsection{Filtered Accuracy via Fact-Checking}

The experiments conducted to evaluate the accuracy of responses generated by the LLaMA 3 70B Instruct model, when filtered through LEAF's fact-checking, are described across five medical datasets. Specifically, we calculate the filtered accuracy, which measures the model’s accuracy on responses that pass the fact-check test (i.e., those with a fact-check score of 1.0). We compare the original accuracy of the model on these datasets with the accuracy filtered through our fact-check system (LEAF) and a baseline fact-check system Factcheck-GPT~\cite{wang2023factcheck}, which uses GPT-3.5 as the rating model with Google Search results as the knowledge source.

\begin{table*}[t!]
\centering
\scalebox{0.8}{
\begin{tabular}{lcccccc}
\toprule
        Dataset & USMLE & MMLU-Medical & PubMedQA & BioASQ & MedMCQA & Average \\
        \midrule
        LLaMA 3 70B Instruct (Original Accuracy) & 73.53 & 85.12 & 60.6 & 80.58 & 71.21 & 74.21\\
        Filtered Accuracy (Factcheck-GPT) & 77.35 & 84.01 & 51.04 & 85.29 & 75.31 & 74.60\\
        Filtered Accuracy (LEAF) & \textbf{86.52} & \textbf{93.01} & \textbf{72.63} & \textbf{96.27} & \textbf{81.62} & \textbf{86.01}\\

        \bottomrule
\end{tabular}
}
\caption{Comparison of original accuracy and filtered accuracy on five medical QA datasets. Filtered accuracy represents the accuracy of LLaMA 3 70B Instruct responses that pass fact-checking (fact-check score = 1) for both Factcheck-GPT~\cite{wang2023factcheck} or LEAF. For LEAF, the fact-check score is calculated as the ratio of supported sentences to the total number of sentences in the response. A score of 1 indicates all sentences in the response are supported by the fact-checking process.}
\label{tab:accuracy_comparison}
\end{table*}


Table \ref{tab:accuracy_comparison} demonstrates the effectiveness of our fact-checking approach in enhancing the accuracy of the LLaMA 3 70B Instruct model’s outputs. As shown, filtering responses using our LEAF fact-check system significantly improves accuracy on all datasets compared to the original model accuracy and the baseline Factcheck-GPT. The highest gains in accuracy are observed when using LEAF, indicating its superior performance in validating factually correct answers. This highlights the robustness of our approach in leveraging fact-checking for validating and improving large language model outputs.

\subsection{Fact-Check-Then-RAG}

To evaluate the effectiveness of our Fact-Check-Then-RAG (FC-RAG) approach, we present the experiments conducted  comparing it to the original performance of the LLaMA 3 70B Instruct model and the standard RAG setting in MedRAG \cite{xiong2024benchmarking}. In MedRAG, the question is used as a query to retrieve relevant documents, which are then included in the prompt. In our FC-RAG approach, we use information obtained in the fact-checking stage to include in the prompt.


\begin{table*}[t!]
\centering
\scalebox{0.8}{
\begin{tabular}{lcccccc}
\toprule
        Dataset & USMLE & MMLU-Medical & PubMedQA & BioASQ & MedMCQA & Average \\
        \midrule
        Llama 3 70B Instruct & 73.53 & 85.12 & 60.60 & 80.58 & 71.21 & 74.21\\
        Llama 3 70B Instruct (MedRAG) & 68.58 & 82.46 & 70.80 & 87.70 & 68.78 & 75.66\\
        Llama 3 70B Instruct (FC-RAG) & \textbf{77.52} & \textbf{86.78} & \textbf{73.60} & \textbf{87.86} & \textbf{72.77} & \textbf{79.71}\\
        \bottomrule
\end{tabular}
}
\caption{Comparison of LLaMA 3 70B original performance, performance when using MedRAG, and FC-RAG on five medical QA datasets. Note that all of model’s parameters remained unchanged.}
\label{tab:factcheck_rag_comparison}
\end{table*}


Table \ref{tab:factcheck_rag_comparison} compares the performance of the Llama 3 70B Instruct model across five medical QA datasets: USMLE, MMLU-Medical, PubMedQA, BioASQ, and MedMCQA. The table lists the original accuracy of the Llama 3 70B Instruct model on each dataset, the accuracy when using MedRAG, and the accuracy using our FC-RAG approach. The LLaMA 3 70B (MedRAG) shows the model's performance when applying the standard RAG method. While MedRAG \cite{xiong2024benchmarking} is designed to improve the model's contextual grounding by providing additional information, the results reveal that it actually harms performance on the USMLE and MMLU-Medical datasets—consistent with findings in MedRAG original paper~\cite{xiong2024benchmarking}. This suggests that while MedRAG can be beneficial in certain contexts, it may introduce noise or irrelevant information in others, leading to decreased accuracy. In contrast, the FC-RAG approach consistently improves accuracy across all datasets. By incorporating fact-checking results into the RAG process, FC-RAG ensures that the model's outputs are more reliable and factually correct. This method leverages verified information during generation, leading to significant performance gains: a 4.99\% improvement on USMLE, 1.66\% on MMLU-Medical, 13.0\% on PubMedQA, 7.28\% on BioASQ, and 1.56\% on MedMCQA compared to the original model performance. These results demonstrate the robustness and efficiency of FC-RAG in enhancing the outputs of large language models, particularly in domains where factual accuracy is critical.

\subsection{Supervised Fine-Tuning on Factually Correct Responses}

In order to assess the effectiveness of a model fine-tuned on fact-checked generated responses, we initiated a series of experiments. The LLaMA 3 8B Instruct model was tested on prompts drawn from five datasets, generating responses that were subsequently fact-checked. We perform supervised fine-tuning on the responses that pass the fact-check test(the response with fact-check score is 1.0). We compare the performance of the SFT model with the original model and also conduct the same experiments on the Factcheck-GPT~\cite{wang2023factcheck}.

\begin{table*}[h!]
\centering
\scalebox{0.8}{
\begin{tabular}{lcccccc}
\toprule
        Dataset & USMLE & MMLU-Medical & PubMedQA & BioASQ & MedMCQA & Average \\
        \midrule
        Llama 3 8B Instruct & 55.46 & 70.98 & 55.20 & 74.27 & 57.78 & 62.74 \\
        Llama 3 8B Instruct(SFT Factcheck-GPT) & 57.03 & 71.99 & 59.60 & 75.40 & 58.71 & 64.55 \\
        Llama 3 8B Instruct(SFT LEAF) & \textbf{60.17} & \textbf{75.85} & \textbf{61.80} & \textbf{78.80} & \textbf{60.75} & \textbf{67.47}\\
        \bottomrule
\end{tabular}
}
\caption{Comparison of original performance, SFT with Factcheck-GPT, and SFT with LEAF on five medical QA datasets.}
    \label{tab:sft_comparison}
\end{table*}


Table \ref{tab:sft_comparison} presents the performance of the Llama 3 8B Instruct model across five medical QA datasets: USMLE, MMLU-Medical, PubMedQA, BioASQ, and MedMCQA. The table lists the original accuracy of the Llama 3 8B Instruct model on each dataset, the accuracy after supervised fine-tuning using Factcheck-GPT, and the accuracy after supervised fine-tuning using our system (LEAF). As evident from the table, SFT on fact-checked responses significantly improves the accuracy of the model on all datasets compared to the original accuracy. Specifically, the SFT approach using LEAF shows notable improvements: an increase of 4.71\% on USMLE, 4.87\% on MMLU-Medical, 6.60\% on PubMedQA, 4.53\% on BioASQ, and 2.97\% on MedMCQA compared to the original model performance. Furthermore, our fact-check system outperforms the baseline, indicating its robustness and efficiency in ensuring the generated answers are factually correct and contextually relevant. This demonstrates the potential of combining fact-checking with fine-tuning to enhance the outputs of large language models.

\begin{table*}[h!]
\centering
\scalebox{0.8}{
\begin{tabular}{lcccccc}
\toprule
        Dataset & USMLE & MMLU-Medical & PubMedQA & BioASQ & MedMCQA & Average \\
        \midrule
        Llama 3 8B Instruct(Lowest ArmoRM score) & 51.92 & 68.69 & 58.40 & 74.60 & 57.54 &62.23\\
        Llama 3 8B Instruct(Highest ArmoRM score) & 56.80 & 73.19 & 60.20 & 78.32 & 59.91 &65.68\\
        $\Delta$(ArmoRM) & 4.88 & 4.50 & 1.80 & 3.72 & 2.37 & 3.45\\
        Llama 3 8B Instruct(Lowest LEAF score) & 48.78 & 68.69 & 53.20 & 73.79 & 55.99 & 60.09\\
        Llama 3 8B Instruct(Highest LEAF score) & 60.33 & 73.55 & 64.60 & 79.94 & 61.42 & 67.97\\
        $\Delta$(LEAF) & \textbf{11.55} & \textbf{4.86} & \textbf{11.40} & \textbf{6.15} & \textbf{5.43} & \textbf{7.88} \\
        \bottomrule
\end{tabular}
}
\caption{Comparison of lowest and highest scored responses using ArmoRM and LEAF across five medical QA datasets. $\Delta$ represents the difference between the highest and lowest performance for each system.}
\label{tab:ranking_comparison}
\end{table*}


\subsection{Fact-Checking as a Ranking Model}

We conducted a series of experiments to assess the effectiveness of our fact-checking system as a ranking model for responses generated by large language models. Five responses were generated using the LLaMA 3 8B Instruct model with a temperature setting of 0.8. Each response was then scored by our fact-checking system, and the performance of the highest and lowest-scored responses was analyzed. For comparison, we also ran similar experiments using ArmoRM.~\cite{wang2024interpretable}, a reward model designed to align LLMs with human preferences. ArmoRM is trained using human preference data, employing a Mixture-of-Experts (MoE) strategy to select suitable reward objectives based on context. Table~\ref{tab:ranking_comparison} presents the performance of the highest and lowest-scored responses using both our fact-checking system and ArmoRM. Additionally, we compute the $\Delta$(difference) between the highest and lowest scores for both systems to facilitate comparison.


\textbf{LLaMA 3 8B (Lowest ArmoRM score):} Performance of the lowest scored response using the ArmoRM reward model.

\textbf{LLaMA 3 8B (Highest ArmoRM score):} Performance of the highest scored response using the ArmoRM reward model.

\textbf{$\Delta$(ArmoRM):} This indicates the difference in performance between the highest and lowest-scored responses using ArmoRM, providing insight into the gap of quality based on the reward model's scoring.

\textbf{LLaMA 3 8B (Lowest LEAF score):} Performance of the lowest scored response using LEAF.

\textbf{LLaMA 3 8B (Highest LEAF score):} Performance of the highest scored response using LEAF.

\textbf{$\Delta$(LEAF):} This indicates the difference in performance between the highest and lowest-scored responses using our fact-checking system. 


As evident from the table, our fact-checking system(LEAF) effectively ranks the responses to highlight the best-performing ones. The larger $\Delta$ values for our system compared to ArmoRM demonstrate the robustness and efficiency of our fact-checking approach in differentiating between high-quality and low-quality responses.

\subsection{SimPO on Ranked Responses}

We design experiments to evaluate the effectiveness of SimPO on responses ranked by our fact-checking system and by ArmoRM \cite{wang2024interpretable}. For each prompt/question, we generate five responses using the Llama 3 8B Instruct model with a temperature setting of 0.8. We then use our fact-checking system and ArmoRM to score these responses, selecting the lowest-scored responses as ``rejected" and the highest-scored responses as ``chosen". We then run SimPO on these chosen and rejected responses.

\begin{table*}[h!]
\centering
\scalebox{0.9}{
\begin{tabular}{lcccccc}
\toprule
        Dataset & USMLE & MMLU-Medical & PubMedQA & BioASQ & MedMCQA & Average \\
        \midrule
        Llama 3 8B Instruct & 55.46 & 70.98 & 55.20 & 74.27 & 57.78 & 62.74\\
        Llama 3 8B Instruct(SimPO ArmoRM) & 56.40 & 72.82 & 59.00 & 76.70 & 59.05 & 64.79\\
        Llama 3 8B Instruct(SimPO LEAF) & \textbf{59.54} & \textbf{73.65} & \textbf{62.00} & \textbf{81.72} & \textbf{60.67} & \textbf{67.52}\\
        \bottomrule
\end{tabular}
}
\caption{Comparison of original performance, SimPO on ArmoRM ranked responses, and SimPO on LEAF ranked responses across five medical QA datasets.}
    \label{tab:simpo_comparison}
\end{table*}

As shown in Table \ref{tab:simpo_comparison}, the SimPO optimization on LEAF-ranked responses results in better performance compared to the optimization on ArmoRM-ranked responses. Specifically, the SimPO approach using LEAF shows significant improvements: an increase of 4.08\% on USMLE, 2.67\% on MMLU-Medical, 6.80\% on PubMedQA, 7.45\% on BioASQ, and 2.89\% on MedMCQA compared to the original model performance. This is attributed to the larger gap between the highest and lowest-scored responses in our fact-checking system, as demonstrated in Table~\ref{tab:ranking_comparison}. A larger gap indicates a more significant distinction between high-quality and low-quality responses, leading to more effective optimization and ultimately better performance after SimPO. The results demonstrate the robustness and efficiency of our fact-checking approach in differentiating between high-quality and low-quality responses, thereby enhancing the performance of the Llama 3 8B Instruct model after training.

\section{Related Work}

\textbf{Evaluating factuality in Model Responses} Evaluating the factuality of model responses is crucial for ensuring the reliability of large language models. Recent studies have demonstrated that LLMs can serve as effective tools for fact verification~\cite{guan-etal-2024-language, tian2023fine}. Improvements in human evaluation techniques have further enhanced factuality assessment~\cite{cheng2024relic}. Factcheck-GPT~\cite{wang2023factcheck} presents an end-to-end solution for annotating factuality in LLM outputs, offering fine-grained labels for verifiability and factual inconsistencies. Inspired by methods that break down responses for evaluation \cite{chern2023factool}, SAFE~\cite{wei2024long} applies a similar approach in the long-form factuality setting, leveraging search-augmented models. Our work adapts SAFE for the medical domain, incorporating question context, deploying Qwen2-72B-Instruct for reliable factuality ratings, and using the MedRAG \cite{xiong2024benchmarking} corpus with ColBERT \cite{khattab2020colbert} retrieval. 


\textbf{Retrieval-Augmented Generation} Retrieval-Augmented Generation, proposed by \cite{yih2020retrieval}, integrates relevant retrieved information into the generation process of LLMs, enhancing their performance on knowledge-intensive tasks. This approach helps improve factualness by grounding the LLMs on provided contexts and supplying up-to-date knowledge that might not be encoded in the models. Many studies have built upon the original RAG framework to further improve its effectiveness, including works by \cite{borgeaud2022improving, ram2023context, gao2023retrieval, jiang2023active}. In the biomedical field, RAG has been explored for literature information-seeking and clinical decision-making \cite{frisoni2022bioreader, naik2022literature, jin2023retrieve}.
However, LLMs often struggle to capture fine-grained knowledge and frequently produce inaccurate or fabricated information, commonly referred to as hallucination. Current RAG methods remain under-explored in the context of fact verification, particularly in terms of accurate evidence retrieval and fine-grained classification. As such, our study introduces the Fact-Check-Then-RAG approach, which integrates a fact-checking stage to further enhance the reliability and accuracy of LLM-generated responses.

\textbf{Learning from Fact-Check via Self-Training} Inspired by the Med-Gemini model's self-training with web search integration to enhance clinical reasoning \cite{saab2024capabilities}, we developed a self-training approach with fact-checking to improve the accuracy and reliability of large language models. Self-training with search involves generating reasoning paths with and without web search, refining the model iteratively by integrating search results and expert demonstrations, which allows the model to be deployed offline on private servers while also improving the efficiency of inference. In contrast, our self-training with fact-checking generates multiple responses to prompts, evaluates them for factual accuracy using a fact-checking system, and fine-tunes the model on validated responses. This method ensures learning from accurate information and reduces hallucinations. While self-training with search relies on real-time web data, our fact-checking approach validates against established knowledge bases, offering a controlled and reliable framework for model enhancement, particularly in low-resource domains.

\section{Conclusion}

In this study, we investigated the potential of fact-checking mechanisms to improve factuality ability in LLM within the context of medical question-answering tasks. 
We validated the effectiveness of our approach through the original model, trained model, Retrieval-Augmented Generation (RAG), and fine-tuning. Through a series of experiments with models such as Llama 3, we demonstrated significant improvements in performance, particularly in enhancing the correctness of generated responses across all datasets compared to traditional RAG, underscoring the effectiveness of leveraging fact-checked information to provide more contextually relevant responses. In our second strategy, the use of fact-checking as a ranking model in conjunction with SimPO further refined the model's output, illustrating a clear path toward higher accuracy and robustness in LLM-generated content. Furthermore, we demonstrate that the system architecture can replace closed-source LLMs integrated with Google Search by using self-deployed open-source LLMs with specialized corpus retrieval. This approach is more controllable and cost-effective, allowing precise tuning for specific datasets and domains while reducing dependency on external APIs.



\bibliography{aaai25}

\begin{thebibliography}{31}
\providecommand{\natexlab}[1]{#1}

\bibitem[{Bang et~al.(2023)Bang, Cahyawijaya, Lee, Dai, Su, Wilie, Lovenia, Ji, Yu, Chung et~al.}]{bang2023multitask}
Bang, Y.; Cahyawijaya, S.; Lee, N.; Dai, W.; Su, D.; Wilie, B.; Lovenia, H.; Ji, Z.; Yu, T.; Chung, W.; et~al. 2023.
\newblock A multitask, multilingual, multimodal evaluation of chatgpt on reasoning, hallucination, and interactivity.
\newblock \emph{arXiv preprint arXiv:2302.04023}.

\bibitem[{Borgeaud et~al.(2022)Borgeaud, Mensch, Hoffmann, Cai, Rutherford, Millican, Van Den~Driessche, Lespiau, Damoc, Clark et~al.}]{borgeaud2022improving}
Borgeaud, S.; Mensch, A.; Hoffmann, J.; Cai, T.; Rutherford, E.; Millican, K.; Van Den~Driessche, G.~B.; Lespiau, J.-B.; Damoc, B.; Clark, A.; et~al. 2022.
\newblock Improving language models by retrieving from trillions of tokens.
\newblock In \emph{International conference on machine learning}, 2206--2240. PMLR.

\bibitem[{Cheng et~al.(2024)Cheng, Zouhar, Arora, Sachan, Strobelt, and El-Assady}]{cheng2024relic}
Cheng, F.; Zouhar, V.; Arora, S.; Sachan, M.; Strobelt, H.; and El-Assady, M. 2024.
\newblock Relic: Investigating large language model responses using self-consistency.
\newblock In \emph{Proceedings of the CHI Conference on Human Factors in Computing Systems}, 1--18.

\bibitem[{Chern et~al.(2023)Chern, Chern, Chen, Yuan, Feng, Zhou, He, Neubig, and Liu}]{chern2023factool}
Chern, I.-C.; Chern, S.; Chen, S.; Yuan, W.; Feng, K.; Zhou, C.; He, J.; Neubig, G.; and Liu, P. 2023.
\newblock Factool: Factuality detection in generative AI-A tool augmented framework for multi-task and multi-domain scenarios. CoRR, abs/2307.13528, 2023. doi: 10.48550.
\newblock \emph{arXiv preprint arXiv.2307.13528}.

\bibitem[{Frisoni et~al.(2022)Frisoni, Mizutani, Moro, and Valgimigli}]{frisoni2022bioreader}
Frisoni, G.; Mizutani, M.; Moro, G.; and Valgimigli, L. 2022.
\newblock Bioreader: a retrieval-enhanced text-to-text transformer for biomedical literature.
\newblock In \emph{Proceedings of the 2022 conference on empirical methods in natural language processing}, 5770--5793.

\bibitem[{Gao et~al.(2023)Gao, Xiong, Gao, Jia, Pan, Bi, Dai, Sun, and Wang}]{gao2023retrieval}
Gao, Y.; Xiong, Y.; Gao, X.; Jia, K.; Pan, J.; Bi, Y.; Dai, Y.; Sun, J.; and Wang, H. 2023.
\newblock Retrieval-Augmented Generation for Large Language Models: A Survey.
\newblock \emph{arXiv e-prints}, arXiv--2312.

\bibitem[{Guan et~al.(2024)Guan, Dodge, Wadden, Huang, and Peng}]{guan-etal-2024-language}
Guan, J.; Dodge, J.; Wadden, D.; Huang, M.; and Peng, H. 2024.
\newblock Language Models Hallucinate, but May Excel at Fact Verification.
\newblock In Duh, K.; Gomez, H.; and Bethard, S., eds., \emph{Proceedings of the 2024 Conference of the North American Chapter of the Association for Computational Linguistics: Human Language Technologies (Volume 1: Long Papers)}, 1090--1111. Mexico City, Mexico: Association for Computational Linguistics.

\bibitem[{Hendrycks et~al.(2020)Hendrycks, Burns, Basart, Zou, Mazeika, Song, and Steinhardt}]{hendrycksmeasuring}
Hendrycks, D.; Burns, C.; Basart, S.; Zou, A.; Mazeika, M.; Song, D.; and Steinhardt, J. 2020.
\newblock Measuring Massive Multitask Language Understanding.
\newblock In \emph{International Conference on Learning Representations}.

\bibitem[{Ji et~al.(2023)Ji, Lee, Frieske, Yu, Su, Xu, Ishii, Bang, Madotto, and Fung}]{ji2023survey}
Ji, Z.; Lee, N.; Frieske, R.; Yu, T.; Su, D.; Xu, Y.; Ishii, E.; Bang, Y.~J.; Madotto, A.; and Fung, P. 2023.
\newblock Survey of hallucination in natural language generation.
\newblock \emph{ACM Computing Surveys}, 55(12): 1--38.

\bibitem[{Jiang et~al.(2023)Jiang, Xu, Gao, Sun, Liu, Dwivedi-Yu, Yang, Callan, and Neubig}]{jiang2023active}
Jiang, Z.; Xu, F.~F.; Gao, L.; Sun, Z.; Liu, Q.; Dwivedi-Yu, J.; Yang, Y.; Callan, J.; and Neubig, G. 2023.
\newblock Active Retrieval Augmented Generation.
\newblock In \emph{Proceedings of the 2023 Conference on Empirical Methods in Natural Language Processing}, 7969--7992.

\bibitem[{Jin et~al.(2021)Jin, Pan, Oufattole, Weng, Fang, and Szolovits}]{jin2021disease}
Jin, D.; Pan, E.; Oufattole, N.; Weng, W.-H.; Fang, H.; and Szolovits, P. 2021.
\newblock What disease does this patient have? a large-scale open domain question answering dataset from medical exams.
\newblock \emph{Applied Sciences}, 11(14): 6421.

\bibitem[{Jin et~al.(2019)Jin, Dhingra, Liu, Cohen, and Lu}]{jin2019pubmedqa}
Jin, Q.; Dhingra, B.; Liu, Z.; Cohen, W.; and Lu, X. 2019.
\newblock PubMedQA: A Dataset for Biomedical Research Question Answering.
\newblock In \emph{Proceedings of the 2019 Conference on Empirical Methods in Natural Language Processing and the 9th International Joint Conference on Natural Language Processing (EMNLP-IJCNLP)}, 2567--2577.

\bibitem[{Jin, Leaman, and Lu(2023)}]{jin2023retrieve}
Jin, Q.; Leaman, R.; and Lu, Z. 2023.
\newblock Retrieve, summarize, and verify: how will ChatGPT affect information seeking from the medical literature?
\newblock \emph{Journal of the American Society of Nephrology}, 34(8): 1302--1304.

\bibitem[{Khattab and Zaharia(2020)}]{khattab2020colbert}
Khattab, O.; and Zaharia, M. 2020.
\newblock Colbert: Efficient and effective passage search via contextualized late interaction over bert.
\newblock In \emph{Proceedings of the 43rd International ACM SIGIR conference on research and development in Information Retrieval}, 39--48.

\bibitem[{Lewis et~al.(2020)Lewis, Perez, Piktus, Petroni, Karpukhin, Goyal, K{\"u}ttler, Lewis, Yih, Rockt{\"a}schel et~al.}]{lewis2020retrieval}
Lewis, P.; Perez, E.; Piktus, A.; Petroni, F.; Karpukhin, V.; Goyal, N.; K{\"u}ttler, H.; Lewis, M.; Yih, W.-t.; Rockt{\"a}schel, T.; et~al. 2020.
\newblock Retrieval-augmented generation for knowledge-intensive nlp tasks.
\newblock \emph{Advances in Neural Information Processing Systems}, 33: 9459--9474.

\bibitem[{Marcus and Davis(2020)}]{marcus2020gpt}
Marcus, G.; and Davis, E. 2020.
\newblock GPT-3, Bloviator: OpenAI’s language generator has no idea what it’s talking about.
\newblock \emph{Technology Review}, 294.

\bibitem[{Meng, Xia, and Chen(2024)}]{meng2024simpo}
Meng, Y.; Xia, M.; and Chen, D. 2024.
\newblock SimPO: Simple Preference Optimization with a Reference-Free Reward.
\newblock \emph{arXiv e-prints}, arXiv--2405.

\bibitem[{Naik et~al.(2022)Naik, Parasa, Feldman, Wang, and Hope}]{naik2022literature}
Naik, A.; Parasa, S.; Feldman, S.; Wang, L.~L.; and Hope, T. 2022.
\newblock Literature-Augmented Clinical Outcome Prediction.
\newblock In \emph{Findings of the Association for Computational Linguistics: NAACL 2022}, 438--453.

\bibitem[{Pal, Umapathi, and Sankarasubbu(2022)}]{pal2022medmcqa}
Pal, A.; Umapathi, L.~K.; and Sankarasubbu, M. 2022.
\newblock Medmcqa: A large-scale multi-subject multi-choice dataset for medical domain question answering.
\newblock In \emph{Conference on health, inference, and learning}, 248--260. PMLR.

\bibitem[{Petroni et~al.(2021)Petroni, Piktus, Fan, Lewis, Yazdani, Cao, Thorne, Jernite, Karpukhin, Maillard et~al.}]{petroni2021kilt}
Petroni, F.; Piktus, A.; Fan, A.; Lewis, P.; Yazdani, M.; Cao, N.; Thorne, J.; Jernite, Y.; Karpukhin, V.; Maillard, J.; et~al. 2021.
\newblock KILT: a Benchmark for Knowledge Intensive Language Tasks.
\newblock In \emph{NAACL-HLT}, 2523--2544. Association for Computational Linguistics.

\bibitem[{Ram et~al.(2023)Ram, Levine, Dalmedigos, Muhlgay, Shashua, Leyton-Brown, and Shoham}]{ram2023context}
Ram, O.; Levine, Y.; Dalmedigos, I.; Muhlgay, D.; Shashua, A.; Leyton-Brown, K.; and Shoham, Y. 2023.
\newblock In-context retrieval-augmented language models.
\newblock \emph{Transactions of the Association for Computational Linguistics}, 11: 1316--1331.

\bibitem[{Saab et~al.(2024)Saab, Tu, Weng, Tanno, Stutz, Wulczyn, Zhang, Strother, Park, Vedadi et~al.}]{saab2024capabilities}
Saab, K.; Tu, T.; Weng, W.-H.; Tanno, R.; Stutz, D.; Wulczyn, E.; Zhang, F.; Strother, T.; Park, C.; Vedadi, E.; et~al. 2024.
\newblock Capabilities of Gemini Models in Medicine.
\newblock \emph{arXiv e-prints}, arXiv--2404.

\bibitem[{Tian et~al.(2023)Tian, Mitchell, Yao, Manning, and Finn}]{tian2023fine}
Tian, K.; Mitchell, E.; Yao, H.; Manning, C.; and Finn, C. 2023.
\newblock Fine-tuning Language Models for Factuality.
\newblock In \emph{NeurIPS 2023 Workshop on Instruction Tuning and Instruction Following}.

\bibitem[{Tsatsaronis et~al.(2015)Tsatsaronis, Balikas, Malakasiotis, Partalas, Zschunke, Alvers, Weissenborn, Krithara, Petridis, Polychronopoulos et~al.}]{tsatsaronis2015overview}
Tsatsaronis, G.; Balikas, G.; Malakasiotis, P.; Partalas, I.; Zschunke, M.; Alvers, M.~R.; Weissenborn, D.; Krithara, A.; Petridis, S.; Polychronopoulos, D.; et~al. 2015.
\newblock An overview of the BIOASQ large-scale biomedical semantic indexing and question answering competition.
\newblock \emph{BMC bioinformatics}, 16: 1--28.

\bibitem[{Wang et~al.(2024)Wang, Xiong, Xie, Zhao, and Zhang}]{wang2024interpretable}
Wang, H.; Xiong, W.; Xie, T.; Zhao, H.; and Zhang, T. 2024.
\newblock Interpretable Preferences via Multi-Objective Reward Modeling and Mixture-of-Experts.
\newblock \emph{arXiv preprint arXiv:2406.12845}.

\bibitem[{Wang et~al.(2023)Wang, Gangi~Reddy, Mujahid, Arora, Rubashevskii, Geng, Afzal, Pan, Borenstein, Pillai et~al.}]{wang2023factcheck}
Wang, Y.; Gangi~Reddy, R.; Mujahid, Z.~M.; Arora, A.; Rubashevskii, A.; Geng, J.; Afzal, O.~M.; Pan, L.; Borenstein, N.; Pillai, A.; et~al. 2023.
\newblock Factcheck-GPT: End-to-End Fine-Grained Document-Level Fact-Checking and Correction of LLM Output.
\newblock \emph{arXiv e-prints}, arXiv--2311.

\bibitem[{Wei et~al.(2024)Wei, Yang, Song, Lu, Hu, Tran, Peng, Liu, Huang, Du et~al.}]{wei2024long}
Wei, J.; Yang, C.; Song, X.; Lu, Y.; Hu, N.; Tran, D.; Peng, D.; Liu, R.; Huang, D.; Du, C.; et~al. 2024.
\newblock Long-form factuality in large language models.
\newblock \emph{arXiv e-prints}, arXiv--2403.

\bibitem[{Xiong et~al.(2024)Xiong, Jin, Lu, and Zhang}]{xiong2024benchmarking}
Xiong, G.; Jin, Q.; Lu, Z.; and Zhang, A. 2024.
\newblock Benchmarking Retrieval-Augmented Generation for Medicine.
\newblock \emph{arXiv e-prints}, arXiv--2402.

\bibitem[{Yang et~al.(2024)Yang, Yang, Hui, Zheng, Yu, Zhou, Li, Li, Liu, Huang et~al.}]{yang2024qwen2}
Yang, A.; Yang, B.; Hui, B.; Zheng, B.; Yu, B.; Zhou, C.; Li, C.; Li, C.; Liu, D.; Huang, F.; et~al. 2024.
\newblock Qwen2 Technical Report.
\newblock \emph{arXiv e-prints}, arXiv--2407.

\bibitem[{Yih(2020)}]{yih2020retrieval}
Yih, S. 2020.
\newblock Retrieval-augmented generation for knowledge-intensive nlp tasks.
\newblock In \emph{Conference on Neural Information Processing Systems, Vancouver, Canada}.

\bibitem[{Zellers et~al.(2019)Zellers, Holtzman, Rashkin, Bisk, Farhadi, Roesner, and Choi}]{zellers2019defending}
Zellers, R.; Holtzman, A.; Rashkin, H.; Bisk, Y.; Farhadi, A.; Roesner, F.; and Choi, Y. 2019.
\newblock Defending against neural fake news.
\newblock \emph{Advances in neural information processing systems}, 32.

\end{thebibliography}

\appendix
\section*{Reproducibility Checklist}

\begin{enumerate}

\item Includes a conceptual outline and/or pseudocode description of AI methods introduced: Yes

\item Clearly delineates statements that are opinions, hypothesis, and speculation from objective facts and results: Yes

\item Provides well-marked pedagogical references for less-familiar readers to gain background necessary to replicate the paper: Yes

\item Does this paper make theoretical contributions? Yes

    \begin{enumerate}
        \item All assumptions and restrictions are stated clearly and formally: Yes
        \item All novel claims are stated formally (e.g., in theorem statements): Yes
        \item Proofs of all novel claims are included: Yes
        \item Proof sketches or intuitions are given for complex and/or novel results: Yes
        \item Appropriate citations to theoretical tools used are given: Yes
        \item All theoretical claims are demonstrated empirically to hold: Yes
        \item All experimental code used to eliminate or disprove claims is included: Yes
    \end{enumerate}

\item Does this paper rely on one or more datasets? Yes

    \begin{enumerate}
        \item A motivation is given for why the experiments are conducted on the selected datasets: Yes
        \item All novel datasets introduced in this paper are included in a data appendix: Yes
        \item All novel datasets introduced in this paper will be made publicly available upon publication of the paper with a license that allows free usage for research purposes: Yes
        \item All datasets drawn from the existing literature (potentially including authors’ own previously published work) are accompanied by appropriate citations: Yes
        \item All datasets drawn from the existing literature (potentially including authors’ own previously published work) are publicly available: Yes
        \item All datasets that are not publicly available are described in detail, with an explanation of why publicly available alternatives are not scientifically satisfying: Yes
    \end{enumerate}

\item Does this paper include computational experiments? Yes

    \begin{enumerate}
        \item Any code required for pre-processing data is included in the appendix: Yes
        \item All source code required for conducting and analyzing the experiments is included in a code appendix: Yes
        \item All source code required for conducting and analyzing the experiments will be made publicly available upon publication of the paper with a license that allows free usage for research purposes: Yes
        \item All source code implementing new methods have comments detailing the implementation, with references to the paper where each step comes from: Yes
        \item If an algorithm depends on randomness, then the method used for setting seeds is described in a way sufficient to allow replication of results: Yes
        \item This paper specifies the computing infrastructure used for running experiments (hardware and software), including GPU/CPU models; amount of memory; operating system; names and versions of relevant software libraries and frameworks: Yes
        \item This paper formally describes evaluation metrics used and explains the motivation for choosing these metrics: Yes
        \item This paper states the number of algorithm runs used to compute each reported result: Yes
        \item Analysis of experiments goes beyond single-dimensional summaries of performance (e.g., average; median) to include measures of variation, confidence, or other distributional information: Yes
        \item The significance of any improvement or decrease in performance is judged using appropriate statistical tests (e.g., Wilcoxon signed-rank): Yes
        \item This paper lists all final (hyper-)parameters used for each model/algorithm in the paper’s experiments: Yes
        \item This paper states the number and range of values tried per (hyper-) parameter during the development of the paper, along with the criterion used for selecting the final parameter setting: Yes
    \end{enumerate}

\end{enumerate}

\appendix
\include{Appendix}
\section{Appendix}
\label{sec:appendix}
\appendix

\subsection{Overview}
This appendix provides supplementary information and detailed examples to support the methodology and results presented in the main paper. It is structured as follows:

\begin{itemize}
    \item \textbf{Datasets:} A comprehensive description of the five medical datasets used in our experiments, including USMLE, MMLU-Medical, PubMedQA, BioASQ, and MedMCQA.
    \item \textbf{Prompts:} Examples of prompts used for query generation, fact-checking, and retrieval-augmented generation, demonstrating how our system interacts with the language models.
    \item \textbf{Fact-Checking Process:} A step-by-step walkthrough of our fact-checking methodology, including:
    \begin{enumerate}
        \item Query generation with context
        \item Retrieval from the MedRAG corpus
        \item Fact-checking with context
    \end{enumerate}
    \item \textbf{Fact-Check-Then-RAG process: } A walkthrough of how to use the fact-checking results to guide the RAG process. 
    \item \textbf{Impact of Fact-Checking and Sample Questions:} An analysis of how fact-checking influences the selection of correct options, illustrated with examples and visualizations. This section includes a set of sample questions from the USMLE-MedQA dataset to demonstrate the system's performance and allow for experiment reproduction.

    \item \textbf{Self-Training Experimental Setup:} Detailed information about the infrastructure, hyperparameters, and training procedures used in our experiments.
    \item \textbf{Limitations:} A discussion of our work's limitations and future improvement. 
\end{itemize}

Each section builds upon the previous ones, providing a comprehensive view of our methodology and its application. The examples and figures throughout the appendix are designed to illustrate key concepts and provide empirical support for our approach.

\subsection{Datasets}

In this subsection, we describe the datasets used in our experiments. We utilize the MIRAGE benchmark \cite{xiong2024benchmarking}, which comprises five medical QA datasets, including three medical examination QA datasets and two biomedical research QA datasets. Specifically, the datasets are as follows:

\textbf{MMLU-Med}~\cite{hendrycksmeasuring}: This dataset includes multiple-choice questions from medical examinations, testing the model's knowledge and reasoning in various medical domains.

\textbf{MedQA-US}~\cite{jin2021disease}: This dataset contains multiple-choice questions from the US medical licensing examination, designed to evaluate the model's understanding of medical concepts and clinical practices.

\textbf{MedMCQA}~\cite{pal2022medmcqa}: This dataset features multiple-choice questions from Indian medical examinations, providing a diverse set of questions that test the model's knowledge in clinical medicine and medical science.

\textbf{PubMedQA*}~\cite{jin2019pubmedqa}: Following the setting in the MIRAGE paper, we use a modified version of PubMedQA where all ground-truth supporting contexts are excluded, resulting in PubMedQA*. This dataset focuses on yes/no questions derived from biomedical research articles, testing the model's ability to answer questions based solely on the questions without additional context.

\textbf{BioASQ-Y/N}~\cite{tsatsaronis2015overview}: This dataset contains yes/no questions from the BioASQ challenge, which aims to test the model's ability to understand and answer questions based on biomedical literature.

We adhere to the same settings as the MIRAGE paper, including only multiple-choice questions related to biomedicine and excluding all ground-truth supporting contexts for the questions. For example, in PubMedQA, we remove the contexts and only use the questions, resulting in PubMedQA*. It is important to note that while we focus on medical QA tasks in this work, our workflow of integrating LLMs with fact-checking is generalizable to any domain and can be applied to various tasks beyond QA. We chose the QA task for its popularity in evaluating LLMs and demonstrating the effectiveness of our proposed workflow.

\subsection{Prompts}
In this section, we provide an overview of the various prompts used in our experiments (Table \ref{tab:prompts}). These prompts were designed to guide the LLM through different stages of processing, including query generation, fact-checking, and retrieval-augmented generation. Each prompt is tailored to specific tasks, ensuring the model receives clear instructions to perform the required actions effectively.
\begin{itemize}
    \item \textbf{\{\_KNOWLEDGE\_PLACEHOLDER\}}: This represents the background information or facts that are provided to the model. It typically includes retrieved documents, or previously established facts that can help the model in its reasoning process. 
    
    \item \textbf{\{\_CONTEXT\_PLACEHOLDER\}}: This contains the specific scenario or question that the model needs to address. In medical QA tasks, this often includes patient information, symptoms, and other relevant details of the case. For example, in USMLE-MedQA, this part is dynamically filled with a question and the corresponding answer options. 
    
    \item \textbf{\{\_STATEMENT\_PLACEHOLDER\}}: This represents a specific claim or assertion that the model needs to evaluate or fact-check based on the given knowledge and context. In our medical QA experiments, this placeholder is filled with individual sentences from the LLM's initial response to a question. Each sentence is fact-checked separately to assess the factual accuracy of the entire response at a granular level.
    
    \item \textbf{\{\_QUESTION\_PLACEHOLDER\}}: In the Fact-Check-Then-RAG prompt, this represents the full question text that the model needs to answer.
    
    \item \textbf{\{\_OPTIONS\_PLACEHOLDER\}}: In the Fact-Check-Then-RAG prompt, this contains the list of multiple-choice options that the model can choose from when answering the question.
\end{itemize}

These placeholders are dynamically filled with appropriate content during the execution of our system, allowing for flexible and context-specific interactions with the language model.

\lstset{
    basicstyle=\ttfamily\tiny,
    columns=fullflexible,
    breaklines=true,
    postbreak=\mbox{\textcolor{red}{$\hookrightarrow$}\space}
}

\begin{table*}[h!]
\centering
\begin{tabular}{|m{2.5cm}|p{13.7cm}|}
\hline
\textbf{Type} & \textbf{Prompt} \\ \hline
Query generation with context & 
\begin{lstlisting}[basicstyle=\small]
Instructions:
1. You have been given a STATEMENT, a CONTEXT and some KNOWLEDGE points.
2. Your goal is to try to find evidence that either supports or does not support the factual accuracy of the given STATEMENT in the given CONTEXT.
3. To do this, you are allowed to issue ONE Google Search query that you think will allow you to find additional useful evidence.
4. Your query should aim to obtain new information that does not appear in the KNOWLEDGE. This new information should be useful for determining the factual accuracy of the given STATEMENT.
5. Format your final query by putting it in a markdown code block.

KNOWLEDGE:
{_KNOWLEDGE_PLACEHOLDER}

CONTEXT:
{_CONTEXT_PLACEHOLDER}

STATEMENT:
{_STATEMENT_PLACEHOLDER}
\end{lstlisting}
\\ \hline
Fact-check with context & 
\begin{lstlisting}[basicstyle=\small]
Instructions:
1. You have been given a STATEMENT, a CONTEXT and some KNOWLEDGE points.
2. Determine whether the given STATEMENT is supported by the given CONTEXT, you can use the given KNOWLEDGE to support your decision if necessary. The STATEMENT is supported if it is a proper action or reasoning given the CONTEXT.
3. Before showing your answer, think step-by-step and show your specific reasoning. 
4. If the STATEMENT is supported by the CONTEXT, be sure to show the supporting evidence.
5. After stating your reasoning, restate the STATEMENT and then determine your final answer based on your reasoning and the STATEMENT.
6. Your final answer should be either "{SUPPORTED_LABEL}" or 
"{NOT_SUPPORTED_LABEL}". Wrap your final answer in square brackets.

KNOWLEDGE:
{_KNOWLEDGE_PLACEHOLDER}

CONTEXT:
{_CONTEXT_PLACEHOLDER}

STATEMENT:
{_STATEMENT_PLACEHOLDER}
\end{lstlisting}
\\ \hline

Fact-Check-Then-RAG & 
\begin{lstlisting}[basicstyle=\small]
Given a multiple choice question, please select the correct answer and also provide a detailed reasoning for your choice. You can using the information provided in the knowledge section if necessary. 

KNOWLEDGE:
{_KNOWLEDGE_PLACEHOLDER}

QUESTION:
{_QUESTION_PLACEHOLDER}

OPTIONS:
{_OPTIONS_PLACEHOLDER}

ANSWER:
\end{lstlisting}
\\ \hline
\end{tabular}
\caption{All prompts used in our work.}
\label{tab:prompts}
\end{table*}

\subsection{Fact-Checking Process}
To evaluate the effectiveness of our fact-checking system, we conducted experiments using the Llama 3 70B Instruct model on several samples of the USMLE-MedQA dataset. For each question, ten responses were generated with a temperature setting of 1.2. These responses were subsequently evaluated using our fact-checking system. The figure \ref{fig:fact_check_results} displays the frequency of each answer option along with the average fact-check score assigned to those options. Notably, the fact-check scores tend to be higher for the correct answers, which are highlighted in gold. This visualization illustrates the correlation between the frequency of selected options and their factual accuracy, as determined by the fact-checking system. The results demonstrate that the fact-checking system can reliably identify and score correct responses, supporting its utility in enhancing the factual accuracy of model outputs.

We present an example from the USMLE dataset to illustrate the fact-checking process. The example involves a 13-year-old boy presenting with severe knee, hip, and groin pain. The prompt for the model was:

\begin{quote}
    \textbf{An example of USMLE Question} \textit{A 13-year-old boy presents to the emergency department with severe knee, hip, and groin pain. The patient has a past medical history notable only for obesity and asthma. His temperature is 98°F (36.7°C), blood pressure is 124/65 mmHg, pulse is 128/min, respirations are 14/min, and oxygen saturation is 99\% on room air. Physical exam is notable for an inability of the patient to bear weight on his left leg and limited range of motion of the left hip. Which of the following is the best management for this patient?}
\end{quote}

\noindent The available choices were:
\begin{itemize}
    \item (A) Casting and crutches
    \item (B) Immobilization of the hip in a Pavlik harness
    \item (C) Supportive therapy and observation
    \item (D) Surgical drainage of the hip
    \item \textbf{(E) Surgical pinning of the femoral head} (Correct)
\end{itemize}

For this prompt, we generated 5 responses using the Llama 3 70B Instruct model with a temperature of 1.2. The responses were then fact-checked, with each sentence in the response being evaluated for factual accuracy against retrieved knowledge. The fact-check score for each response was calculated as the ratio of sentences supported by the retrieved knowledge to the total number of sentences in the response.

Table \ref{tab:fact_check_example} illustrates the LLM original generated responses, and their selected options, corresponding fact-check scores.
In the markup text, sentences that were \textbf{not supported} by the retrieved knowledge are highlighted in \textcolor{red}{red}, while sentences that were supported remain in black.

We will take the first response in Table \ref{tab:fact_check_example} as an example to show how to do fact-check with context.
\begin{quote}
    \textbf{An Example of LLM original response} \textit{\textcolor{red}{**(D) Surgical drainage of the hip** **Reasoning:**} This patient's symptoms and physical exam findings are highly suggestive of a septic hip, also known as infectious arthritis of the hip. The patient's severe pain, limited range of motion, and inability to bear weight on the left leg are all consistent with an inflammatory response in the hip joint. The fact that the patient is afebrile (normal temperature) does not rule out septic arthritis, as some patients may not develop a fever. The past medical history of obesity and asthma are not directly related to the patient's current symptoms. The patient's vital signs indicate tachycardia (elevated heart rate), which may be due to pain or inflammation. 	\textcolor{red}{Given the high likelihood of septic arthritis, the best management for this patient is surgical drainage of the hip (option D).} Delaying drainage can lead to permanent damage to the joint and surrounding bone. Surgical drainage will allow for removal of infected fluid and tissue, as well as collection of a sample for microbiological analysis to guide antibiotic therapy. The other options are not appropriate in this scenario: * (A) Casting and crutches are not indicated for a potentially infected joint. * (B) Pavlik harness is a device used to treat developmental dysplasia of the hip in infants, not for septic arthritis in adolescents. * (C) Supportive therapy and observation would not adequately address the potential infection and may lead to further harm. * (E) Surgical pinning of the femoral head is not a treatment for septic arthritis; it may be used for fractures or other orthopedic conditions, but not for an infected joint. 	\textcolor{red}{Therefore, the best management for this patient is surgical drainage of the hip (option D).}}
\end{quote}

\begin{table*}[htbp]
    \centering
    \resizebox{\textwidth}{!}{
    \begin{tabular}{|c|c|p{17cm}|}
        \hline
        \textbf{Selected Option} & \textbf{Fact-Check Score} & \textbf{Response} \\
        \hline
        \textbf{D} & 0.75 & \textcolor{red}{**(D) Surgical drainage of the hip**} 	\textcolor{red}{**Reasoning:**} This patient's symptoms and physical exam findings are highly suggestive of a septic hip, also known as infectious arthritis of the hip. The patient's severe pain, limited range of motion, and inability to bear weight on the left leg are all consistent with an inflammatory response in the hip joint. The fact that the patient is afebrile (normal temperature) does not rule out septic arthritis, as some patients may not develop a fever. The past medical history of obesity and asthma are not directly related to the patient's current symptoms. The patient's vital signs indicate tachycardia (elevated heart rate), which may be due to pain or inflammation. 	\textcolor{red}{Given the high likelihood of septic arthritis, the best management for this patient is surgical drainage of the hip (option D).} Delaying drainage can lead to permanent damage to the joint and surrounding bone. Surgical drainage will allow for removal of infected fluid and tissue, as well as collection of a sample for microbiological analysis to guide antibiotic therapy. The other options are not appropriate in this scenario: * (A) Casting and crutches are not indicated for a potentially infected joint. * (B) Pavlik harness is a device used to treat developmental dysplasia of the hip in infants, not for septic arthritis in adolescents. * (C) Supportive therapy and observation would not adequately address the potential infection and may lead to further harm. * (E) Surgical pinning of the femoral head is not a treatment for septic arthritis; it may be used for fractures or other orthopedic conditions, but not for an infected joint. 	\textcolor{red}{Therefore, the best management for this patient is surgical drainage of the hip (option D).}  \\
        \hline
        \textbf{B} & 0.5 & 	\textcolor{red}{**Answer**: (B) Immobilization of the hip in a Pavlik harness} 	\textcolor{red}{**Reasoning**: Based on the symptoms and physical exam findings, the most likely diagnosis is a slipped capital femoral epiphysis (SCFE).} SCFE is a common condition affecting adolescents, particularly obese individuals, where the epiphysis (growth plate) at the proximal end of the femur slips, causing hip pain and limited mobility. The patient's age, symptoms, and physical exam findings are consistent with SCFE. The inability to bear weight on the left leg and limited range of motion of the left hip suggest a hip-related issue. SCFE is more common in obese individuals, and the patient's history of obesity supports this diagnosis. 	\textcolor{red}{Immobilization of the hip in a Pavlik harness (option B) is the best management for this patient.} 	\textcolor{red}{The Pavlik harness is a device that maintains the hip in a position that reduces the risk of further slippage and allows for gentle reduction of the epiphysis.} 	\textcolor{red}{This approach is often used for SCFE, especially in cases where the slippage is mild to moderate.} \textcolor{red}{The other options are not appropriate for this patient:} * Casting and crutches (option A) would not be effective in reducing the risk of further slippage or promoting healing. * Supportive therapy and observation (option C) would not provide adequate stability or treatment for the condition. 	\textcolor{red}{* Surgical drainage of the hip (option D) is not indicated, as SCFE is not typically associated with an infectious process.} * Surgical pinning of the femoral head (option E) may be necessary in some cases of SCFE, but it is not the initial management of choice. Surgery is often reserved for more severe cases or when orthopedic management fails. 	\textcolor{red}{In conclusion, immobilization of the hip in a Pavlik harness is the best management for this patient, given the likely diagnosis of SCFE and the need to reduce the risk of further slippage and promote healing.}  \\
        \hline
        \textbf{E} & 1.0 & (E) Surgical pinning of the femoral head **Reasoning**: The patient's symptoms, particularly the sudden onset of severe knee, hip, and groin pain, along with the inability to bear weight and limited range of motion of the left hip, suggest a diagnosis of slipped capital femoral epiphysis (SCFE). SCFE is a common condition in adolescents, particularly in those who are obese, which is consistent with the patient's past medical history. SCFE occurs when the femoral head slips out of place at the growth plate, leading to a mismatch between the head and neck of the femur. This can cause significant pain, stiffness, and impaired mobility in the affected hip and surrounding areas. Now, let's evaluate the options: (A) Casting and crutches are typically used for fractures or sprains, which are not likely diagnoses based on the patient's symptoms. (B) Immobilization in a Pavlik harness is usually used for developmental dysplasia of the hip (DDH) in infants, not for SCFE in adolescents. (C) Supportive therapy and observation may be appropriate for minor injuries or self-limiting conditions, but SCFE requires more prompt and definitive treatment to prevent long-term complications, such as osteoarthritis and hip deformities. (D) Surgical drainage of the hip is not typically indicated for SCFE, as there is no clear indication of infection or abscess formation. (E) Surgical pinning of the femoral head is the most appropriate management for this patient. This procedure involves stabilizing the femoral head with one or more pins to prevent further slippage and promote healing. Early surgical intervention can help prevent long-term complications and improve outcomes for patients with SCFE. Therefore, the correct answer is (E) Surgical pinning of the femoral head.  \\
        \hline
        \textbf{D} & 0.64 & \textcolor{red}{**Answer**: D) Surgical drainage of the hip} 	\textcolor{red}{**Reasoning**: Based on the patient's symptoms and physical exam findings, the most likely diagnosis is a septic hip, which is a medical emergency.} The patient's severe pain, inability to bear weight, and limited range of motion of the left hip suggest an inflammatory process in the hip joint. Although the patient's temperature is normal, the patient's pulse is elevated, which could indicate an underlying infection. \textcolor{red}{Given the patient's symptoms and physical exam findings, surgical drainage of the hip is the best management option.} Septic arthritis of the hip joint requires urgent surgical intervention to drain the infected joint fluid, wash out the joint, and administer antibiotics to prevent long-term damage to the joint. Option A, casting and crutches, is not appropriate for a suspected septic hip, as it would not address the underlying infection. Option B, immobilization of the hip in a Pavlik harness, is used to treat developmental dysplasia of the hip in infants, not septic hip in adolescents. Option C, supportive therapy and observation, would not be sufficient to manage a septic hip, as it would not address the underlying infection and could lead to long-term joint damage. 	\textcolor{red}{Option E, surgical pinning of the femoral head, is not indicated in this scenario, as it is used to treat femoral neck fractures or slipped capital femoral epiphyses, not septic hip.} In conclusion, surgical drainage of the hip is the best management option for this patient to ensure prompt treatment of the suspected septic hip and prevent long-term joint damage.  \\
        \hline
        \textbf{D} & 0.73 & \textcolor{red}{The correct answer is (D) Surgical drainage of the hip.} Here's a detailed reasoning for my choice: Based on the patient's symptoms and physical exam findings, the most likely diagnosis is septic arthritis of the left hip. The patient presents with severe pain in the knee, hip, and groin area, along with an inability to bear weight on the left leg and limited range of motion of the left hip. \textcolor{red}{These symptoms are consistent with an infectious process affecting the joint.} The patient's past medical history of obesity and asthma is not directly related to the current presentation, but it's worth noting that obesity may be a contributing factor to the development of septic arthritis. The vital signs show a slightly elevated pulse rate, which may indicate some degree of distress or pain, but the patient's temperature is normal, which is unusual for septic arthritis. However, this does not rule out the diagnosis, as some patients may not develop fever, especially early in the course of the infection. The correct management for septic arthritis is prompt surgical drainage of the infected joint to remove the infected fluid and relieve pressure on the surrounding tissues. This is essential to prevent long-term damage to the joint and surrounding bones. \textcolor{red}{Option (A) Casting and crutches is not appropriate, as immobilization alone will not address the underlying infection.} Option (B) Immobilization of the hip in a Pavlik harness is also not correct, as this type of orthosis is typically used for developmental dysplasia of the hip (DDH) in infants, not for septic arthritis. Option (C) Supportive therapy and observation is not sufficient for septic arthritis, as this condition requires urgent surgical intervention to prevent complications. Option (E) Surgical pinning of the femoral head may be necessary in some cases of hip disorders, but it is not the primary treatment for septic arthritis. \textcolor{red}{Therefore, the best management for this patient is surgical drainage of the hip (Option D) to treat the underlying infection and prevent long-term consequences.}  \\
        \hline
    \end{tabular}}
            \caption{Generated responses, selected options, fact-check scores, and response generated for the example question.}
    \label{tab:fact_check_example}
\end{table*}

\subsubsection{Query Generation with Context}
First, we use the "Query generation with context" prompt (as shown in Table \ref{tab:prompts}) to generate a search query. For this example, the CONTEXT is the full question text, and the STATEMENT is a response generated by the LLM. A sample STATEMENT from the above response could be:

\begin{quote}
    \textit{Given the high likelihood of septic arthritis, the best management for this patient is surgical drainage of the hip (option D).}
\end{quote}

The LLM then generates a search query based on this information:

\begin{lstlisting}[basicstyle=\small]
13-year-old boy knee hip groin pain unable to bear weight best management
\end{lstlisting}

The detail is shown in Figure \ref{fig:prompt_query_generation}.

\subsubsection{Retrieval from MedRAG Corpus}
Next, we use this query to retrieve 3 relevant documents from the MedRAG corpus using ColBERT retrieval. The retrieved documents provide additional context for fact-checking. This process is as shown in Figure \ref{fig:query_to_medrag}.

\subsubsection{Fact-Check with Context}
Finally, we use the "Fact-check with context" prompt (Table \ref{tab:prompts}) to evaluate the factual accuracy of the STATEMENT. The KNOWLEDGE placeholder is filled with the retrieved documents, and the CONTEXT and STATEMENT remain the same as in the query generation step. The filled prompt is shown in Figure \ref{fig:prompt_fact_check_within_context}.

The LLM then evaluates the statement and provides a reasoning:

\begin{quote}
    \textit{The STATEMENT is not supported by the CONTEXT. The CONTEXT does not provide any information that would support a diagnosis of septic arthritis or the need for surgical drainage of the hip. The patient's symptoms are more consistent with conditions such as slipped capital femoral epiphysis (SCFE), which is common in obese adolescents and typically requires surgical pinning of the femoral head.}
    
    \textit{Final answer: [Not Supported]}
\end{quote}

This process is repeated for each sentence in the LLM's response, allowing for a granular assessment of the response's factual accuracy. The overall fact-check score for the response is then calculated as the ratio of supported sentences to the total number of sentences.

The results demonstrate that our fact-checking system can reliably identify and score correct responses, supporting its utility in enhancing the factual accuracy of model outputs.

\begin{figure*}[htbp]
\begin{tcolorbox}[
colback=white!10!white,
colframe=black!75!black,
title=An Example Prompt for Query Generation with Context]
Instructions:\\
1. You have been given a STATEMENT, a CONTEXT and some KNOWLEDGE points. \\
2. Your goal is to try to find evidence that either supports or does not support the factual accuracy of the given STATEMENT in the given CONTEXT. \\
3. To do this, you are allowed to issue ONE Google Search query that you think will allow you to find additional useful evidence. \\
4. Your query should aim to obtain new information that does not appear in the KNOWLEDGE. This new information should be useful for determining the factual accuracy of the given STATEMENT. \\
5. Format your final query by putting it in a markdown code block. \\
KNOWLEDGE: \{\_KNOWLEDGE\_PLACEHOLDER\}\\
CONTEXT: A 13-year-old boy presents to the emergency department with severe knee, hip, and groin pain. The patient has a past medical history notable only for obesity and asthma. His temperature is 98°F (36.7°C), blood pressure is 124/65 mmHg, pulse is 128/min, respirations 14/min, and oxygen saturation is 99\% on room air. Physical exam is notable for an inability of the patient to bear weight on his left leg and limited range of motion of the left hip.  Which of the following is the best management for this patient. The available choices were: (A) Casting and crutches (B) Immobilization of the hip in a Pavlik harness (C) Supportive therapy and observation (D) Surgical drainage of the hip (E) Surgical pinning of the femoral head. \\
STATEMENT: Given the high likelihood of septic arthritis, the best management for this patient is surgical drainage of the hip (option D).\\
\noindent\rule{\linewidth}{0.4pt}

\noindent\textbf{LLM Response:}\\
\texttt{13-year-old boy knee hip groin pain unable to bear weight best management}
\end{tcolorbox}
    \caption{An Example Prompt for Query Generation with Context}
    \label{fig:prompt_query_generation}
\end{figure*}

\begin{figure*}[htbp]
\begin{tcolorbox}[
colback=white!10!white,
colframe=black!75!black,
title=An Example Query to MedRAG Corpus]
13-year-old boy knee hip groin pain unable to bear weight best management\\
\noindent\rule{\linewidth}{0.4pt}

\noindent\textbf{Retrieved Docs from MedRAG Corpus with ColBERT Retrieval:}\\
\texttt{in extreme cases and in older children.Slipped Capital Femoral EpiphysisChildren ages 10 to 16 years can develop displacement of the epiphysis on the femoral neck with no history of injury. The slippage occurs through the weak zone (hypertrophic zone) of the growth plate. When slippage occurs in young patients, check for endocrine disorders such as hypothyroid-ism, renal osteodystrophy, and growth hormone deficiency. Slipped capital femoral epiphysis (SCFE) is associated with African-American heritage and obesity, and it is more com-mon in boys than in girls. One-quarter of cases are bilateral. In patients with endocrine etiology, the condition is usually bilateral. Patients generally present with groin and anterior thigh pain, and the patient may have antalgic gait and a limp. Patient may present with knee pain that can lead to missing the diagnosis. In pediatric patients with knee pain, the ipsi-lateral hip should be assessed as well.Examination of the patient will show obligatory\\ 
The mean age at onset is 6 years, with a range of 3 to 8 years. It is twice as common in male children. The patient or family will describe an acute onset of pain in the groin/hip, anterior thigh, or knee. Irritation of the obturator nerve can cause referred pain in the thigh and knee when the pathology is at the hip. Patients with transient synovitis are often afebrile, walk with a painful limp, and have normal to minimally elevated white blood cell count, C-reactive protein, and erythrocyte sedimentation rate compared with bacterial diseases of the hip (Table 199-1). Table 197-3 lists the differential diagnosis of a limping child. Anteroposterior and frog-leg radiographs of the hip are usually normal. Ultrasonography may reveal a joint effusion. It is mandatory to rule out septic arthritis in the presence of effusion with a joint aspiration and cell count.\\ 
and pelvic osteoto-mies, are done in older age groups and in more severe cases. Osteonecrosis of the femoral head is a possible complication of treatment and can result in pain and decreased range of motion.Legg-Calvé-Perthes DiseaseOsteonecrosis of the proximal femoral epiphysis can cause flattening of the femoral head called Legg-Calvé Perthes disease. The age at presentation is between 4 and 8 years of age and occurs more in males, usually affecting one side. Younger age at presentation (less than 6 years old) will have a better prognosis. The patient presents with groin or knee pain, decreased hip motion, and a limp. Treatment includes traction, physical therapy, abduction exercises, and crutches. Restoration of range of motion is important. Femoral and pelvic osteotomies may be needed in extreme cases and in older children.Slipped Capital Femoral EpiphysisChildren ages 10 to 16 years can develop displacement of the epiphysis on the femoral neck with no history of injury.}
\end{tcolorbox}
    \caption{An example query to MedRAG Corpus and 3 retrieved documents}
    \label{fig:query_to_medrag}
\end{figure*}

\subsection{Fact-Check-Then-RAG}
After the initial fact-checking process, if the LLM's response is found to contain inaccuracies, we employ the Fact-Check-Then-RAG approach to improve the response. This method leverages the knowledge retrieved during the fact-checking stage to generate a more accurate answer.

Using our example question about the 13-year-old boy, let's walk through the Fact-Check-Then-RAG process:

First, we use the "Fact-Check-then-RAG" prompt (as shown in Table \ref{tab:prompts}). The KNOWLEDGE placeholder is filled with the relevant information retrieved during the fact-checking process. For our example, this might include:

\begin{quote}
\textit{Slipped capital femoral epiphysis (SCFE) is associated with African-American heritage and obesity, and it is more common in boys than in girls. Patients generally present with groin and anterior thigh pain, and the patient may have antalgic gait and a limp. Patient may present with knee pain that can lead to missing the diagnosis. In pediatric patients with knee pain, the ipsilateral hip should be assessed as well.}
\end{quote}

The QUESTION placeholder contains the original question text, and the OPTIONS placeholder lists the available choices. The prompt for the LLM would then look like Figure \ref{fig:prompt_fc_rag}.

The LLM then generates a new response based on this prompt. It excludes the option D based on the knowledge retrieved from previous fact-checking, and reaches the correct answer: 
\begin{quote}
\textit{(D) Surgical drainage of the hip is not typically indicated for SCFE, as there is no clear indication of infection or abscess formation.}\\
... \\
\textit{Therefore, the correct answer is (E) Surgical pinning of the femoral head.}
\end{quote}

This Fact-Check-Then-RAG process allows the LLM to generate a more accurate and well-reasoned response by incorporating the relevant medical knowledge retrieved during the fact-checking stage. The resulting answer is not only correct but also provides a detailed explanation grounded in factual information.

\subsection{Impact of Fact-Checking and Sample Questions}

\begin{figure*}[h]
    \centering
    \includegraphics[width=0.7\textwidth]{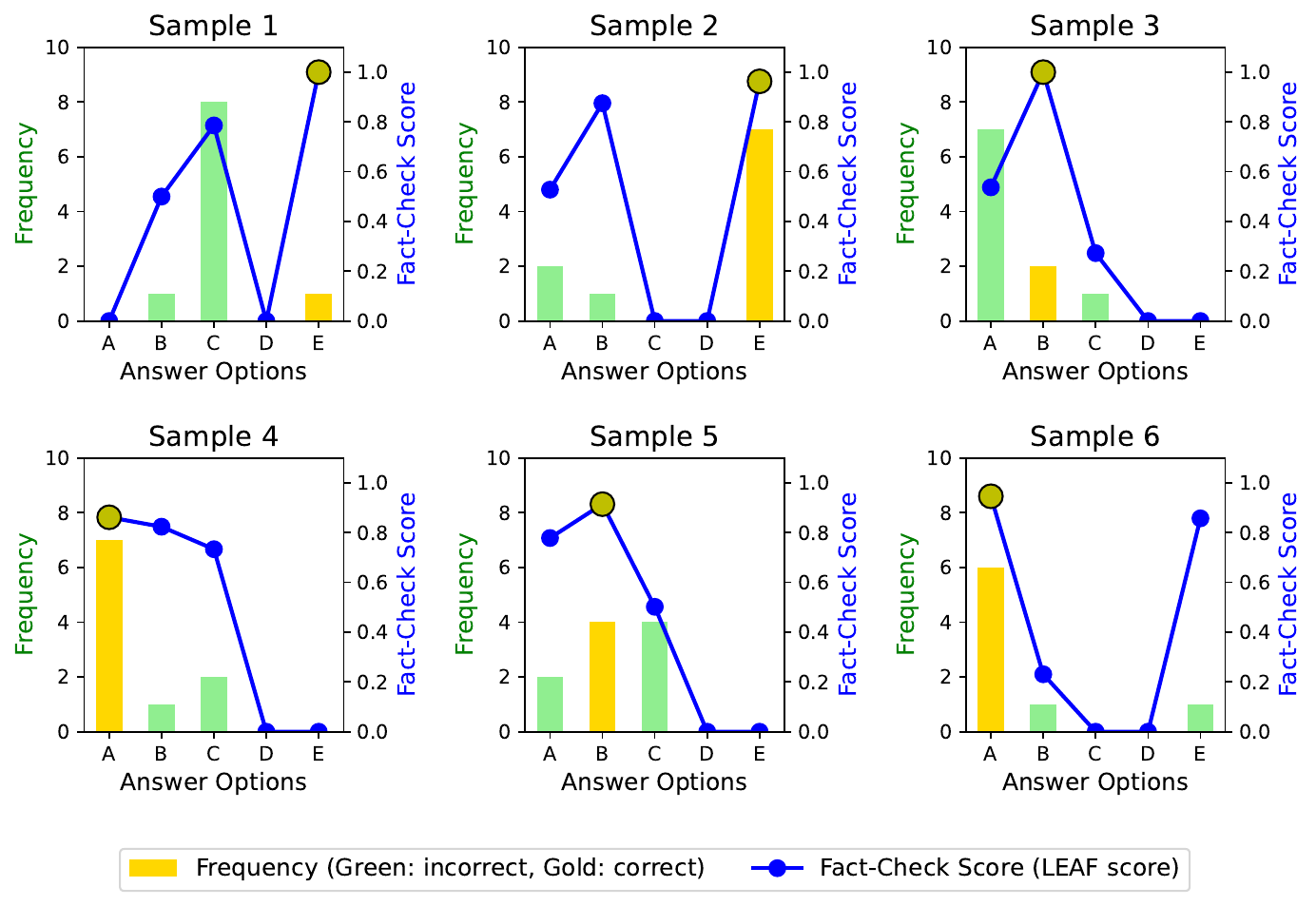}
    \caption{Frequency of selected answer options and their corresponding fact-check scores across multiple samples from the USMLE-MedQA dataset using the Llama 3 70B Instruct model. For each question, ten responses were generated with a temperature setting of 1.2. The fact-check system assigned scores to each option, with higher scores indicating higher factual accuracy. The correct answers, highlighted in gold, consistently received higher fact-check scores. }
    \label{fig:fact_check_results}
\end{figure*}

To demonstrate the effectiveness of our fact-checking system, we conducted experiments using the Llama 3 70B Instruct model on multiple samples from the USMLE-MedQA dataset. Figure \ref{fig:fact_check_results} illustrates the results of these experiments, showing the frequency of selected answer options and their corresponding fact-check scores.

For each of the six sample questions, we generated ten responses using a temperature setting of 1.2. Our fact-checking system then evaluated these responses, assigning scores to each option. The results reveal several key insights:

\textbf{Correlation with Correct Answers:} Across all samples, the correct answers (highlighted in gold) consistently received higher fact-check scores. This strong correlation demonstrates the ability of our fact-checking system to identify factually accurate responses.

\textbf{Handling of Ambiguity:} In some cases, such as sample 4, multiple options received relatively high fact-check scores. This suggests that our system can capture nuanced differences in factual accuracy, even when multiple options may have some degree of correctness.

\textbf{Consistency Across Samples:} The pattern of higher fact-check scores for correct answers is consistent across all six samples, indicating the robustness of our approach across different types of medical questions.

\textbf{Potential for Improving Model Performance:} The clear distinction in fact-check scores between correct and incorrect answers suggests that our system could be effectively used to enhance the model's decision-making process, potentially improving its overall performance on medical QA tasks.

To provide context for these results, we present the six sample questions from the USMLE-MedQA dataset used in this analysis, shown in Figure \ref{fig:sample-questions-1-3} and Figure \ref{fig:sample-questions-4-6}.

These sample questions cover a range of medical scenarios and concepts, demonstrating the versatility of our fact-checking system across different types of medical knowledge and reasoning tasks.

\subsection{Self-Training Experimental Setup}

\textbf{Optimization with SimPO} The second part of our self-training approach utilizes Simple Preference Optimization \cite{meng2024simpo} to rank and optimize responses based on their factual accuracy. SimPO aligns the reward formulation directly with the generation metric, eliminating the need for a reference model. This process involves Fact-Check as Ranking Model: 

\begin{itemize}
    \item Fact-Check as Ranking Model: The fact-checking system assigns scores to generated responses based on their factual accuracy. The highest-scoring responses are selected as ``chosen" and the lowest-scoring as ``rejected."
    \item SimPO Objective: The SimPO objective is designed to maximize the difference in rewards between the chosen and rejected responses. The reward is calculated as:
    \begin{equation}
        r_{SimPO}(x, y) = \frac{\beta}{|y|} \sum_{i=1}^{|y|} \log \pi_{\theta}(y_i | x, y_{<i})
    \end{equation}
    where $\beta$ is a scaling constant. 

    \item Target Reward Margin: Additionally, we introduce a target reward margin term, $\gamma > 0$, to the Bradley-Terry objective to ensure that the reward for the winning response, $r(x, y_w)$, exceeds the reward for the losing response, $r(x, y_l)$, by at least $\gamma$:
    \begin{equation}
        p(y_w \succ y_l | x) = \sigma (r(x, y_w) - r(x, y_l) - \gamma).
    \end{equation}
    
    Finally, we obtain the SimPO objective by incorporating the length-normalized reward:
    \begin{align}
    L_{SimPO}(\pi_{\theta}) = & -\mathbb{E}_{(x,y_w,y_l) \sim D} \nonumber \\
    & \Bigg[\log \sigma \Bigg( \frac{\beta}{|y_w|} \log \pi_{\theta}(y_w|x) \nonumber \\
    & - \frac{\beta}{|y_l|} \log \pi_{\theta}(y_l|x) - \gamma \Bigg) \Bigg].
    \end{align}

\end{itemize}

\subsubsection{Hyperparameters for Training}

The training of the LLaMA 3 8B Instruct model was carefully configured using a set of hyperparameters designed to optimize the model's performance on the selected tasks. The key hyperparameters and their settings are summarized in Table \ref{tab:hyperparameters}.

The learning rate was set to \(1.0 \times 10^{-6}\), a value selected after initial experimentation to balance the rate of convergence with the stability of training. A batch size of 4 per device was chosen to ensure that the model could effectively utilize the available GPU memory, while the gradient accumulation steps were set to 8 to allow for a larger effective batch size without exceeding memory limits.

The maximum sequence length was set to 2048 tokens, with a prompt length of 1800 tokens, ensuring that the model could process lengthy inputs and generate comprehensive responses. The AdamW optimizer was selected for its effectiveness in handling weight decay during training, and the cosine learning rate scheduler was used to gradually reduce the learning rate, facilitating smoother convergence.

The warmup ratio of 0.1 was implemented to gently ramp up the learning rate at the beginning of training, reducing the risk of instability in the early stages. The number of training epochs was set to 5, balancing training time with the need for thorough model training.

Specific to SimPO, the beta and gamma hyperparameters were set to 2.5 and 1.4, respectively. These values were selected based on prior research and experimentation, optimizing the model’s preference ordering during training. Finally, a seed of 42 was used to ensure reproducibility of the results.

\begin{table}[h!]
    \centering
    \begin{tabular}{|l|l|}
        \hline
        \textbf{Hyperparameter} & \textbf{Value} \\
        \hline
        \textbf{Learning Rate} & 1.0e-6 \\
        \textbf{Batch Size per Device} & 4 \\
        \textbf{Gradient Accumulation Steps} & 8 \\
        \textbf{Max Sequence Length} & 2048 \\
        \textbf{Max Prompt Length} & 1800 \\
        \textbf{Optimizer} & AdamW \\
        \textbf{LR Scheduler Type} & Cosine \\
        \textbf{Warmup Ratio} & 0.1 \\
        \textbf{Number of Training Epochs} & 5 \\
        \textbf{Beta (SimPO)} & 2.5 \\
        \textbf{Gamma (SimPO)} & 1.4 \\
        \textbf{Seed} & 42 \\
        \hline
    \end{tabular}
    \caption{Summary of Hyperparameters for Training with SimPO.}
    \label{tab:hyperparameters}
\end{table}

\subsubsection{Infrastructure}

All experiments presented in this paper were conducted using a computing environment equipped with four NVIDIA H100 80GB GPUs. These GPUs are built on the Hopper architecture and feature HBM3 memory, providing exceptional performance for large-scale AI and machine learning tasks.

This high-performance hardware configuration enabled efficient handling of the computationally intensive tasks required for training and evaluating large language models across multiple medical datasets.

\subsubsection{Self-Training Experiments}

In this set of experiments, we focused on evaluating the impact of self-training using the Llama 3 8B Instruct model across five medical datasets. The process began by generating five responses for each prompt, with each prompt corresponding to a question in the selected medical datasets: USMLE, MMLU-Medical, PubMedQA, BioASQ, and MedMCQA.

After generating the responses, we applied two different approaches for each dataset: 
\begin{itemize}
    \item Supervised Fine-Tuning on Fact-Checked Responses: In this approach, we fine-tuned the model using only the responses that passed a rigorous fact-checking process. This ensured that the model learned from the most accurate data available.
    \item Simple Preference Optimization with Fact-Check Ranking: Here, we utilized fact-check scores to rank the generated responses. The highest-ranked responses were used for further optimization of the model via SimPO, refining the model’s output quality based on factual correctness.
\end{itemize}

Each of these self-training methods—SFT and SimPO—was performed separately on each dataset to assess their individual impact on the model's performance. After the training process, we evaluated the accuracy and reliability of the fine-tuned models across the same medical QA datasets, allowing us to determine the effectiveness of each self-training approach. 

It is important to note that all fine-tuning in this experiment was conducted as full fine-tuning without the use of any LoRA (Low-Rank Adaptation) techniques.

\begin{figure*}[htbp]
\begin{tcolorbox}[
colback=white!10!white,
colframe=black!75!black,
title=An Example Prompt for Fact-Check with Context]
Instructions:\\
1. You have been given a STATEMENT, a CONTEXT and some KNOWLEDGE points.\\
2. Determine whether the given STATEMENT is supported by the given CONTEXT, you can use the given KNOWLEDGE to support your decision if necessary. The STATEMENT is supported if it is a proper action or reasoning given the CONTEXT.\\
3. Before showing your answer, think step-by-step and show your specific reasoning. \\
4. If the STATEMENT is supported by the CONTEXT, be sure to show the supporting evidence.\\
5. After stating your reasoning, restate the STATEMENT and then determine your final answer based on your reasoning and the STATEMENT. \\
6. Your final answer should be either "\{SUPPORTED\_LABEL\}" or "\{NOT\_SUPPORTED\_LABEL\}". Wrap your final answer in square brackets.

KNOWLEDGE: \\
(1). in extreme cases and in older children ... \\
(2). The mean age at onset is 6 years, with a range of 3 to 8 years... \\
(3). and pelvic osteoto-mies, are done in older age groups and in more severe cases... \\
(4). ... \\

CONTEXT: A 13-year-old boy presents to the emergency department with severe knee, hip, and groin pain. The patient has a past medical history notable only for obesity and asthma. His temperature is 98°F (36.7°C), blood pressure is 124/65 mmHg, pulse is 128/min, respirations 14/min, and oxygen saturation is 99\% on room air. Physical exam is notable for an inability of the patient to bear weight on his left leg and limited range of motion of the left hip.  Which of the following is the best management for this patient. The available choices were: (A) Casting and crutches (B) Immobilization of the hip in a Pavlik harness (C) Supportive therapy and observation (D) Surgical drainage of the hip (E) Surgical pinning of the femoral head. \\
STATEMENT: Given the high likelihood of septic arthritis, the best management for this patient is surgical drainage of the hip (option D).\\
\noindent\rule{\linewidth}{0.4pt}

\noindent\textbf{LLM Response:}\\
\texttt{The given STATEMENT is "Given the high likelihood of septic arthritis, the best management for this patient is Surgical drainage of the hip."\\To determine if this STATEMENT is supported by the CONTEXT, I will analyze the information provided in the CONTEXT and compare it with the STATEMENT.\\ The CONTEXT describes a 13-year-old boy who presents with severe knee, hip, and groin pain. The patient has a past medical history of obesity and asthma. The physical exam is notable for an inability to bear weight on his left leg and limited range of motion of the left hip. The CONTEXT does not provide any information about the cause of the patient\'s symptoms.\\ The STATEMENT suggests that the best management for this patient is Surgical drainage of the hip. However, the CONTEXT does not provide any information that would support a diagnosis of a condition that would require surgical drainage of the hip. The CONTEXT does not mention any signs of inflammation or infection, which would be necessary to consider surgical drainage of the hip.\\ Furthermore, the KNOWLEDGE points provide information about conditions that may cause hip pain in children, such as transient synovitis, slipped capital femoral epiphysis (SCFE), and Legg-Calve-Perthes disease. None of these conditions would be managed with surgical drainage of the hip without evidence of a joint effusion or septic arthritis.\\ Therefore, the STATEMENT is not supported by the CONTEXT. The CONTEXT does not provide any information that would support the need for surgical drainage of the hip.\\ Final answer: [Not Supported]}
\end{tcolorbox}
    \caption{An example prompt for Fact-Check with context. The final answer to the statement is [Not Supported].}
    \label{fig:prompt_fact_check_within_context}
\end{figure*}

\begin{figure*}[htbp]
\begin{tcolorbox}[
colback=white!10!white,
colframe=black!75!black,
title=An Example Prompt for Fact-Check-Then-RAG]
Given a multiple choice question, please select the correct answer and also provide a detailed reasoning for your choice. You can using the information provided in the knowledge section if necessary. \\

KNOWLEDGE:\\
(1). in extreme cases and in older children ... \\
(2). The mean age at onset is 6 years, with a range of 3 to 8 years... \\
(3). and pelvic osteoto-mies, are done in older age groups and in more severe cases... \\
(4). ... \\

QUESTION:
A 13-year-old boy presents to the emergency department with severe knee, hip, and groin pain. The patient has a past medical history notable only for obesity and asthma. His temperature is 98°F (36.7°C), blood pressure is 124/65 mmHg, pulse is 128/min, respirations 14/min, and oxygen saturation is 99\% on room air. Physical exam is notable for an inability of the patient to bear weight on his left leg and limited range of motion of the left hip.  Which of the following is the best management for this patient. \\

OPTIONS: (A) Casting and crutches (B) Immobilization of the hip in a Pavlik harness (C) Supportive therapy and observation (D) Surgical drainage of the hip (E) Surgical pinning of the femoral head. \\

ANSWER: \\
\noindent\rule{\linewidth}{0.4pt}

\noindent\textbf{LLM Response:}\\
\texttt{(E) Surgical pinning of the femoral head\\ **Reasoning**: \\ The patient's symptoms, particularly the sudden onset of severe knee, hip, and groin pain, along with the inability to bear weight and limited range of motion of the left hip, suggest a diagnosis of slipped capital femoral epiphysis (SCFE). SCFE is a common condition in adolescents, particularly in those who are obese, which is consistent with the patient's past medical history. \\ SCFE occurs when the femoral head slips out of place at the growth plate, leading to a mismatch between the head and neck of the femur. This can cause significant pain, stiffness, and impaired mobility in the affected hip and surrounding areas.\\ Now, let's evaluate the options:\\(A) Casting and crutches are typically used for fractures or sprains, which are not likely diagnoses based on the patient's symptoms. \\ (B) Immobilization in a Pavlik harness is usually used for developmental dysplasia of the hip (DDH) in infants, not for SCFE in adolescents.\\ (C) Supportive therapy and observation may be appropriate for minor injuries or self-limiting conditions, but SCFE requires more prompt and definitive treatment to prevent long-term complications, such as osteoarthritis and hip deformities.\\ (D) Surgical drainage of the hip is not typically indicated for SCFE, as there is no clear indication of infection or abscess formation.\\ (E) Surgical pinning of the femoral head is the most appropriate management for this patient. This procedure involves stabilizing the femoral head with one or more pins to prevent further slippage and promote healing. Early surgical intervention can help prevent long-term complications and improve outcomes for patients with SCFE.\\ Therefore, the correct answer is (E) Surgical pinning of the femoral head.}
\end{tcolorbox}
    \caption{An example prompt for Fact-Check-Then-RAG}
    \label{fig:prompt_fc_rag}
\end{figure*}

\section{Limitations}

Despite the promising results, our study has several limitations that need to be addressed in future work. One significant limitation is the speed and computational efficiency of the fact-checking system. The current implementation requires multiple iterations of inference with LLMs and several retrieval operations for each sentence in the responses. This process can be time-consuming and computationally intensive, potentially limiting the scalability and real-time applicability of our approach. 

Additionally, our study primarily focused on the medical domain, leveraging datasets and corpora specific to healthcare. While this domain specificity ensured relevance and precision, it also limits the generalizability of our findings to other fields. Extending our approach to diverse domains and evaluating its effectiveness across various types of knowledge-intensive tasks will be crucial for broader applicability.

Our future works will also explore LEAF's performance upper bounds by leveraging more comprehensive medical corpora and investigating the impact of multiple rounds of self-training. Additionally, we plan to integrate stronger fact-checking models, such as Meta's LLaMA 405B, to enhance the precision of our fact-verification process and extend LEAF's applicability to other knowledge-intensive domains beyond healthcare.

\begin{figure*}[htbp]
\centering
\begin{minipage}{\textwidth}
\footnotesize
\textbf{Sample 1:} A 13-year-old boy presents to the emergency department with severe knee, hip, and groin pain. The patient has a past medical history notable only for obesity and asthma. His temperature is 98°F (36.7°C), blood pressure is 124/65 mmHg, pulse is 128/min, respirations are 14/min, and oxygen saturation is 99\% on room air. Physical exam is notable for an inability of the patient to bear weight on his left leg and limited range of motion of the left hip. Which of the following is the best management for this patient?

\textbf{Choices:}
\begin{enumerate}[label=(\Alph*)]
\item Casting and crutches
\item Immobilization of the hip in a Pavlik harness
\item Supportive therapy and observation
\item Surgical drainage of the hip
\item Surgical pinning of the femoral head
\end{enumerate}

\textbf{Sample 2:} A 36-year-old nursing home worker presents to the clinic with the complaints of breathlessness, cough, and night sweats for the past 2 months. She further expresses her concerns about the possibility of contracting tuberculosis as one of the patients under her care is being treated for tuberculosis. A PPD skin test is done and reads 11 mm on day 3. Chest X-ray demonstrates a cavitary lesion in the right upper lobe. The standard anti-tuberculosis medication regimen is started. At a follow-up appointment 3 months later the patient presents with fatigue. She has also been experiencing occasional dizziness, weakness, and numbness in her feet. Physical exam is positive for conjunctival pallor. Lab work is significant for a hemoglobin level of 10 g/dL and mean corpuscular volume of 68 fl. What is the most likely cause of her current symptoms?

\textbf{Choices:}
\begin{enumerate}[label=(\Alph*)]
\item Decreased methionine synthesis
\item Inhibition of ferrochelatase
\item Increased homocysteine degradation
\item Increased GABA production
\item Decreased ALA synthesis
\end{enumerate}

\textbf{Sample 3:} A 72-year-old woman is admitted to the hospital for treatment of unstable angina. Cardiac catheterization shows occlusion that has caused a 50\% reduction in the diameter of the left circumflex artery. Resistance to blood flow in this vessel has increased by what factor relative to a vessel with no occlusion?

\textbf{Choices:}
\begin{enumerate}[label=(\Alph*)]
\item 64
\item 16
\item 8
\item 4
\item 32
\end{enumerate}
\end{minipage}
\caption{Sample questions 1-3 from the USMLE-MedQA dataset}
\label{fig:sample-questions-1-3}
\end{figure*}

\begin{figure*}[htbp]
\centering
\begin{minipage}{\textwidth}
\footnotesize
\textbf{Sample 4:} A 49-year-old woman is brought to the emergency department with progressive dyspnea and cough which she developed approx. 8 hours ago. 2 weeks ago she had a prophylactic ovariectomy because of a family history of ovarian cancer. She is known to have type 2 diabetes mellitus and stage 1 hypertension, but she does not take her antihypertensives because she is not concerned about her blood pressure. Also, she has a history of opioid abuse. She takes metformin 1000 mg and aspirin 81 mg. She has been smoking 1 pack of cigarettes per day for 22 years. Her vital signs are as follows: blood pressure 155/80 mm Hg, heart rate 101/min, respiratory rate 31/min, and temperature 37.9C (100.2F). Blood saturation on room air is 89\%. On examination, the patient is dyspneic and acrocyanotic. Lung auscultation reveals bilateral rales over the lower lobes. A cardiac examination is significant for S2 accentuation best heard in the second intercostal space at the left sternal border and S3 presence. There is no leg edema. Neurological examination is within normal limits. Arterial blood gases analysis shows the following results:
pH 7.49
PaO2 58 mm Hg
PaCO2 30 mm Hg
HCO3-  22 mEq/L
Based on the given data, which of the following could cause respiratory failure in this patient?

\textbf{Choices:}
\begin{enumerate}[label=(\Alph*)]
\item Increased alveolar dead space due to absent perfusion of certain alveoli
\item Ischemia of the medullary respiratory center neurons
\item Alveolar fibrosis
\item Depression of the respiratory center via opioid receptors activation
\item Decreased V/Q due to bronchial obstruction
\end{enumerate}

\textbf{Sample 5:} While in the ICU, a 62-year-old male undergoes placement of a Swan-Ganz catheter to evaluate his right heart pressures. All pressures are found to be within normal limits, and the cardiology fellow records a pulmonary wedge pressure of 10 mmHg. Which of the following are normal values for the pressures that will be obtained from this patient's right ventricle?

\textbf{Choices:}
\begin{enumerate}[label=(\Alph*)]
\item 25/10 mmHg
\item 25/5 mmHg
\item 10/0 mmHg
\item 100/5 mmHg
\item 100/70 mmHg
\end{enumerate}

\textbf{Sample 6:} A previously healthy 6-year-old boy is brought to the physician because of generalized malaise and a palpable swelling in the left axilla. The parents report that 2 weeks ago, his daycare group visited an animal shelter, after which he developed a rash on the left hand. His temperature is 38.5°C (101.3°F). Physical examination shows three linear crusts on an erythematous background on the dorsum of the left hand. There is tender left-sided axillary and cervical lymphadenopathy. Histopathologic examination of an axillary lymph node shows necrotizing granulomas. The most likely causal organism of this patient's clinical findings is also involved in the pathogenesis of which of the following conditions?

\textbf{Choices:}
\begin{enumerate}[label=(\Alph*)]
\item Bacillary angiomatosis
\item Burkitt lymphoma
\item Condylomata lata
\item Brucellosis
\item Bubonic plague
\end{enumerate}
\end{minipage}
\caption{Sample questions 4-6 from the USMLE-MedQA dataset}
\label{fig:sample-questions-4-6}
\end{figure*}

\end{document}